\documentclass{article}

\PassOptionsToPackage{numbers, compress}{natbib}
 \usepackage[main,final]{neurips_2026}

\usepackage[utf8]{inputenc} %
\usepackage[T1]{fontenc}    %
\usepackage{graphicx}       %
\usepackage{hyperref}       %
\usepackage{url}            %
\usepackage{booktabs}       %
\usepackage{diagbox}         %
\usepackage{amsfonts}       %
\usepackage{amsmath}        %
\usepackage{amssymb}        %
\usepackage{nicefrac}       %
\usepackage{microtype}      %
\usepackage[table]{xcolor}  %
\usepackage{algorithm}      %
\usepackage{algorithmic}   %
\usepackage{wrapfig}
\usepackage{ifthen}
\usepackage{multirow}      %
\usepackage{siunitx}        %
\usepackage{tikz}
\usepackage{cleveref}
\usepackage{rotating}
\definecolor{gainGreen}{RGB}{20,125,65}
\definecolor{bgpurple}{RGB}{170, 112, 161}

\definecolor{lightbluehl}{RGB}{225,240,255}
\definecolor{lightorangehl}{RGB}{255,230,213}
\newcommand{\hlblue}[1]{%
  \tikz[baseline=(X.base)] 
    \node[fill=lightbluehl, rounded corners=1pt, inner xsep=2pt, inner ysep=0.3pt] (X) {#1};%
}
\newcommand{\hlorange}[1]{%
  \tikz[baseline=(X.base)] 
    \node[fill=lightorangehl, rounded corners=1pt, inner xsep=2pt, inner ysep=0.3pt] (X) {#1};%
}

\title{Toward Better Geometric Representations for Molecule Generative Models}

\newcommand{\modelname}{LENSEs}

\author{%
Shaoheng Yan$^{1,2,*}$\\
\And
Zian Li$^{1,3,*}$
\And
Cai Zhou$^{5}$
\And
Qiaojing Huang$^{2,4}$
\AND
Kai Liu$^{2,4}$
\And
Muhan Zhang$^{1,6,\dagger}$
\AND
\quad\\[-1em]
$^1$Institute for Artificial Intelligence, PKU,\quad $^2$ByteDance AI Drug Discovery\\
$^3$School of Intelligence Science and Technology, PKU,\quad $^4$Anew Labs\\
$^5$Department of Electrical Engineering and Computer Science, MIT\\
$^6$State Key Laboratory of General Artificial Intelligence, PKU
}
\usepackage{etoc}
\usepackage{titletoc}
\newcommand{\pmerr}[2]{#1$_{\pm \text{#2}}$}
\usepackage{tikz}
\usepackage{bm}
\usepackage{makecell}
\begin{document}

\maketitle

\let\thefootnote\relax\footnotetext{$^*$ Equally contributed.}
\let\thefootnote\relax\footnotetext{$^\dagger$ Corresponding to Muhan Zhang {<muhan@pku.edu.cn>}, Kai Liu {<liukai0824@bytedance.com>} and Qiaojing Huang {<huangqiaojing@bytedance.com>}.}

\begin{abstract}
Geometric representation-conditioned generation decouples molecule representation modeling from 3D structure generation: a molecule-level representation is generated first, and a 3D molecule is then generated conditioned on it, which significantly improves both efficiency and quality. However, its effectiveness is fundamentally bounded by the quality of the representation space: representations from pretrained molecular encoders such as UniMol are non-smooth, suffer from dimensional collapse, and are not fully exploited during generative training. In this work, we propose \modelname{}, a framework that better exploits molecule representations in representation-conditioned generation. \modelname{} introduces three complementary mechanisms: (1) a \emph{representation head}, trained jointly with the generator, that aggregates multi-level features from the pretrained encoder; (2) a \emph{molecule perceptual loss} that optimizes the generator in a semantically informative representation space; and (3) a \emph{node-level representation alignment (REPA) loss} that aligns the generator's hidden states with encoder representations, narrowing the semantic gap between pretraining and generation. On the challenging GEOM-DRUG benchmark, \modelname{} achieves 97.28\% validity and 98.51\% molecule stability, surpassing existing advanced methods. Beyond generation quality, the refined representations are substantially richer: the effective rank increases by 1.35$\times$ on GEOM-DRUG. With the encoder and representation head frozen, the refined representations also outperform the raw encoder features on \emph{all} QM9 properties and MoleculeNet datasets, establishing generative training with alignment objectives as a promising \emph{pretraining paradigm} for molecular encoders.
\end{abstract}

\section{Introduction}
\label{sec:intro}
3D molecule generation has emerged as a cornerstone of modern drug design~\citep{zeng2022deep,cheng2021molecular,tang2024survey}, evolving from language-model-based string generation (\citep{weininger1988smiles,krenn2020self,cheng2023group}) toward direct 3D conformation generation~\citep{hoogeboom2022equivariant, xu2022geodiff}. As the molecules being modeled grow increasingly large and complex, generation methods must capture fine-grained geometric and topological features with high precision~\citep{adak2025molvision,wang2024x2}. Inspired by effective methods in image domains such as Diffusion~\citep{ho2020denoising, song2020denoising} and Flow Matching~\citep{lipman2022flow}, advanced generative methods have been proposed for molecule generation~\citep{ watson2023novo,corso2022diffdock,walters2020applications,tom2024self,xue2019advances,dunn2026flowmol3,zeng2026propmolflow,reidenbach2026applications,poletukhin20263d,grisoni2020bidirectional,xu2023geometric}.

A promising recent paradigm for molecular generation is \emph{geometric representation-conditioned generation} (GeoRCG)~\citep{li2024geometric}, which decouples molecular representation modeling from generator training. Instead of directly generating a 3D molecule with an equivariant backbone~\citep{hoogeboom2022equivariant}, GeoRCG first samples a molecular representation, i.e., a molecule-level vector, using a Representation Diffusion Model (RDM)~\citep{yang2023diffusion,xu2023geometric}, and then generates the 3D molecular structure conditioned on this representation. The representation space is defined by a pretrained encoder, such as UniMol~\citep{unimol} or Frad~\citep{frad}. During training, the encoder extracts molecular representations, which are then used to train both the RDM and the representation-conditioned molecular generator in a self-conditioning framework~\citep{li2024return}. As shown in~\cite{li2024geometric}, this design substantially improves both sample quality and generation efficiency.

However, the success of this paradigm is fundamentally bounded by the quality of the molecule representations~\citep{li2024geometric,yan2025georecon,yin2025multi}. 
We identify two concrete manifestations of this bottleneck. First,
\textbf{a collapsed and non-smooth representation manifold}. Current
pretrained molecular encoders concentrate their molecule-level
representations in a \hlblue{low-dimensional subspace} (\Cref{tab:lipschitz}). Such a degenerate geometry leaves little room for the RDM to
place distinguishable molecules, and small conformational perturbations
can still induce disproportionately large latent shifts, forcing prior
work~[25] to inject \emph{hand-tuned noise} during training as a compensatory
mechanism. Moreover, due to the lack of \hlorange{generative pretraining}, these
encoder are optimized primarily for property prediction
rather than generation, which may lack enough semantic guidance for
generation tasks.

Second, \textbf{underexploited multi-level structural semantics}. We observe that pretrained molecular encoders exhibit hierarchical features: shallow layers capture local structural motifs such as functional groups, while deeper layers encode more global molecular semantics, analogously to hierarchical representations in computer vision~\citep{gatys2016image,lin2017feature}. However, existing representation-conditioned generation pipelines use the encoder mainly as a condition provider, without explicitly transferring this hierarchical structured knowledge to the generator. As a result, the generator must relearn semantics from scratch, increasing optimization difficulty and cost.

Motivated by these observations, we develop a mechanism that allows the representation space to be optimized during molecule generator training, so that it can guide molecular generation in a smoother and higher-quality manner, while also transferring the encoder's hierarchical molecular semantics to the generator for better training. To this end, we propose \textbf{L}atent \textbf{E}nhancement for \textbf{N}on-smooth \textbf{S}tructural \textbf{E}ncoding\textbf{s} (\textbf{LENSEs}), a framework that systematically improves the conditioning latent space for geometric representation-conditioned molecule generation. Rather than passively relying on a frozen pretrained encoder, \modelname{} explicitly refines the representation space during generator training through three mechanisms: (1) a representation head that maps the encoder’s multi-layer features into a variationally regularized latent space to improve smoothness and sampleability; (2) a molecule-specific perceptual loss~\citep{johnson2016perceptual} that preserves multi-scale semantic information in the encoder feature space, encouraging the generated molecules to match meaningful structural patterns; and (3) a representation alignment (REPA) loss~\citep{yu2024repa} that directly aligns the generator’s internal representations with the encoder’s higher-level features, thereby reducing the generator’s learning burden and accelerating training.

We validate our approach through complementary experiments:
(i) \modelname{} improves over the direct GeoRCG baseline~\citep{li2024geometric} on GEOM-DRUG generation and outperforms recent competitive approaches such as CanonFlow~\citep{zhou2026rethinking}, establishing state-of-the-art performance in both structural and physical plausibility metrics;
(ii) Lipschitz constant analysis reveals that the refined representation space is significantly smoother than that of the raw encoder;
(iii) downstream QM9 regression with the refined representation outperforms the pretrained encoder output, demonstrating that our generative training pipeline also serves as an effective pretraining paradigm.

\section{Related Work}

\textbf{3D Molecule Generation with Equivariant Models.}\quad
Generating 3D molecules requires models that respect geometric symmetries. EDM \cite{hoogeboom2022equivariant} established a key paradigm by jointly modeling atom coordinates and atom types with an equivariant diffusion framework. GCDM \cite{s42004_024_01233_z} improved structural plausibility by incorporating local geometric features such as bond angles and torsions, while MiDi \cite{midi} improved chemical validity through mixed discrete-continuous diffusion. To improve efficiency, EquiFM \cite{2312_07168v1} replaced diffusion simulation with flow matching~\citep{lipman2022flow}, and geometric optimal transport methods \cite{hong2024accelerating} further optimized transport paths. Other structure-based drug design methods include TargetDiff \cite{guan20233d} and DiffSBDD \cite{schneuing2024structure}, which generate molecules conditioned on 3D protein pockets. 

\textbf{Latent Diffusion and Representation-Conditioned Generation.}\quad
To avoid the high cost of direct coordinate-space generation, latent diffusion methods~\citep{rombach2022high} generate molecules in compressed latent spaces. GeoLDM~\citep{2305_01140v1} pioneered this idea by learning an equivariant autoencoder that maps molecular geometries into continuous latent codes, reducing dimensionality while preserving equivariance. GeoRCG~\citep{li2024geometric} further introduced representation-conditioned generation, where a pretrained geometric encoder provides property-rich representations, a lightweight diffusion model captures their distribution, and a generator reconstructs structures conditioned on them. Despite their efficiency, these methods share a key limitation: they assume the encoder latent space is naturally suitable for generation. However, encoders trained for property prediction often induce semantically disorganized and non-smooth latent spaces, leaving a fundamental bottleneck that prior work has mainly treated with ad hoc fixes such as noise injection.

\textbf{Perceptual Losses and Representation Alignment.}\quad
In the image domain, perceptual loss is often preferred over low-level pixel-wise objectives because it aligns more closely with human perception~\citep{zhang2018unreasonable}. It has been shown to be effective in a range of optimization settings~\citep{rombach2022high,song2023consistency,mirzaei2023spin}, where directly optimizing pixel loss can lead to undesirable blurriness.
REPA~\citep{yu2024repa} and REED~\citep{wang2025learning} further brought this idea to image diffusion by aligning diffusion hidden states with a pretrained vision encoder. In molecular generation, geometry-complete methods~\citep{s42004_024_01233_z} suggested that richer structural supervision also improves plausibility, while REED~\citep{wang2025learning} aligned the internal features of molecular diffusion models with external molecule-level pretrained representations to accelerate and improve generative training.
Our work systematically leverages representations for both \emph{conditioning} and \emph{alignment} to improve 3D molecular generation, establishing a bidirectional relationship between generation quality and representation quality.

\section{LENSEs: Latent Enhancement for Non-smooth Structural Encodings}
\label{sec:motivation}

\subsection{Observations}

\textbf{Degenerate latent geometry of the pretrained encoder.} \quad Current
pretrained encoders are primarily trained on coordinate denoising
tasks~\citep{zaidi2022pre,frad,unimol}. Specifically, for a given molecule, random noise is added to its 3D coordinates, which are then input to the encoder, whose objective is to denoise the molecule and recover its original stable conformation. As shown in~\citep{zaidi2022pre}, this paradigm is analogous to learning a physical force field, with denoising corresponding to the conservative case.

Despite its downstream effectiveness, atom-level pretraining can yield pooled molecular representations with limited effective dimensionality. On GEOM-DRUG, raw UniMol representations exhibit an effective rank of only 65.32 (\Cref{tab:lipschitz}) and tightly concentrated intermolecular distances (\Cref{fig:appendix-pairwise}), suggesting limited variation along many latent directions. This degenerate geometry leaves the RDM with little room to
place distinguishable molecules apart. The same representations are also
locally unstable---small coordinate perturbations can induce large shifts in
the latent space, as reflected by the large Lipschitz constants in
\Cref{tab:lipschitz}, a phenomenon also observed
in~\cite{yan2025georecon,perez2026self}. Both properties are detrimental to
representation-conditioned generative tasks. To mitigate this issue,
GeoRCG~\citep{li2024geometric} improves the representation--molecule
connection learned by the conditional generator by adding random,
hand-crafted channel-uniform noise to the encoder's raw representations
during training.

\begin{figure}[t]
  \centering
  \includegraphics[width=\textwidth]{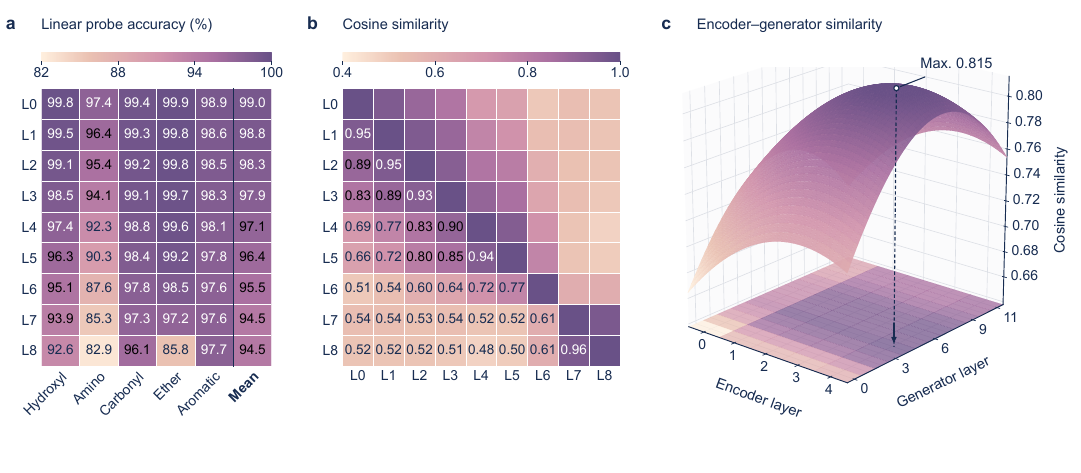}
  \vspace{-3em}
\caption{\textbf{Layer-wise representation analysis.} (a) Linear probe accuracy for selected functional groups and the overall mean across nine Frad encoder layers on QM9. (b) Cross-layer cosine similarity of encoder representations. (c) Node-level cosine similarity between generator and encoder layers, with projections onto the layer plane. Darker colors indicate higher values within each panel.}
  
  \label{fig:layer-analysis}
  \vspace{-1em}
\end{figure}

\textbf{Semantic Information in Shallow Encoder Layers.} \quad The GeoRCG framework leverages the pretrained encoder only as a representation-space definer, limiting its role to providing molecule-level representations. However, as shown in~\cite{zhang2018unreasonable, yu2024repa}, the intermediate features of deep pretrained networks can capture richer semantic information, providing stronger semantic similarity between entities and aligning well with the generator’s learned intermediate features.

To investigate this in the molecular domain, we conduct an additional probing experiment to examine whether shallow encoder layers preserve meaningful chemical semantics.
Specifically, we use Frad~\citep{frad} to perform a systematic layer-wise probe analysis on 30,000 randomly selected QM9~\citep{ramakrishnan2014quantum} molecules.
We train a linear probe on the frozen representation of each layer to classify ten common molecule functional groups, including hydroxyl, amino, and carbonyl. The goal is to assess whether such information is linearly accessible from each layer's representation. 

As shown in \Cref{fig:layer-analysis}a, with full results in \Cref{tab:layer-probe-full}, first-layer probes achieve the highest classification accuracy across all ten functional groups, with gaps up to 14\% for groups such as amino and ether.
To understand this trend, we visualize layer-wise latent similarity in \Cref{fig:layer-analysis}b, which reveals a clear two-block structure: the first six layers are mutually similar, while the last two form a separate block.
This suggests that final-layer conditioning may discard semantic signals preserved in earlier layers.

This observation is further supported by the learned layer weights used for generation: instead of assigning all weight to the final-layer representation, the model allocates nearly 20\% of the total weight to earlier encoder layers.
Together, these results show that early encoder layers encode motif semantics that should be preserved and exploited during generation.

\textbf{Alignment between Encoder and Generator Features.} \quad Beyond encoder-side analysis, we further observe from the public checkpoint of prior work~\citep{li2024geometric} that the layer-wise representations of the molecular generator are highly similar to those of the pretrained encoder. The detailed grid-wise comparison is shown in \Cref{fig:layer-analysis}c. Intuitively, this indicates that the two models share common structural patterns and encoded knowledge. 
The similarity between representations of generators and perceptual models has also been observed in computer vision~\citep{yu2024repa}. 
This suggests that the generator is not learning from scratch, but instead tends to recover part of the encoder's representational structure during training. Motivated by this observation, we explicitly align the representations of the pretrained encoder and the generator through a projector, thereby accelerating and improving the generator's acquisition of chemically meaningful structural knowledge.

\subsection{Unlocking Potentials of Molecule Representations}
\label{sec:method}

\begin{figure}
    \centering
    \includegraphics[width=\textwidth]{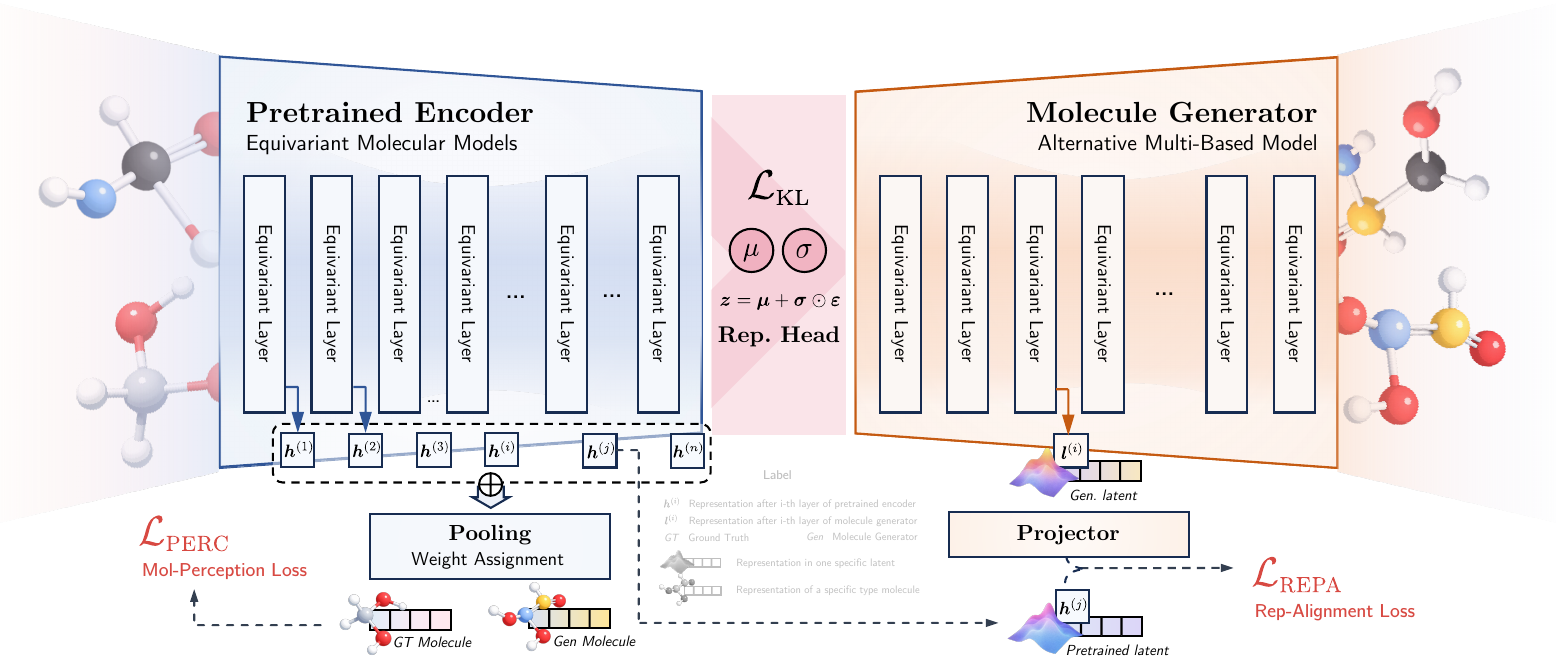}
    \vspace{-2pt}
    \caption{\textbf{The framework of \modelname{}.} \textbf{Phase I:} the molecule generator and the representation head are trained based on a pretrained encoder. The representation head takes as input a learnable pooling of representations from different encoder layers, while the molecule generator performs conditional generation using the reparameterized latent produced by the representation head. \textbf{Phase II:} the RDM is trained to model the latent distribution of target molecules in the mean space of the representation head, thereby serving as a representation generator. \textbf{Phase III:} molecule generation is carried out using the trained RDM together with the molecule generator.}
    \label{fig:method}
    \vspace{-2pt}
\end{figure}
In this section, we present {\modelname{}}, an enhanced framework for  equivariant molecule generation conditioned on geometric representations. It introduces three components: (1) a representation head over the pretrained encoder latent space to \hlblue{improve representation quality}; (2) a perceptual loss ($\mathcal{L}_{\textsc{PERC}}$) for semantic supervision in encoder space; and (3) a representation projection alignment loss ($\mathcal{L}_{\textsc{repa}}$) to align generator and encoder features.

\textbf{Notations.}\quad A 3D molecule is represented as $\mathcal{M} = (\bm{x}, \bm{a}, \mathcal{E})$, where $\bm{x} \in \mathbb{R}^{N \times 3}$ denotes atomic coordinates, $\bm{a}$ encodes atom types, and $\mathcal{E}$ encodes chemical bonds. The goal is to learn an equivariant generative distribution $p_\theta(\mathcal{M})$ satisfying SE(3)-equivariance. Given a molecule $\mathcal{M}$, a pretrained encoder $f_{\text{enc}}$ extracts a geometric representation:
\begin{equation}
    \bm{h} = f_{\text{enc}}(\mathcal M) \in \mathbb{R}^{D}.
\end{equation}

\textbf{Representation Space Trained with Generative Tasks.} \quad The raw output of the pretrained encoder is a deterministic function of the input molecule, and its latent geometry is not explicitly optimized during generative training. To obtain a more structured and generation-friendly latent space, we introduce a representation head on top of the multi-layer encoder representations, enabling adaptive aggregation of different level semantics. 
Specifically, let $\bm{H}^{(l)}(\mathcal M) \in \mathbb{R}^{d}$ denote the representation of molecule $\mathcal M$ extracted from the $l$-th encoder layer, where $l=1,\ldots,L$ and $L$ is the number of encoder layers. To exploit information from different representation depths, we aggregate these layer-wise graph representations through learnable layer pooling:
\begin{equation}
    \bm{g}(\mathcal M) = \sum_{l=1}^{L}\alpha_l \bm{H}^{(l)}(\mathcal M) , \qquad
    \alpha_l = \textsc{Softmax}(\bm{w})_l,
\end{equation}
where $\bm{g}(\mathcal M)  \in \mathbb{R}^{d}$ is the pooled graph representation, $\bm{w}\in\mathbb{R}^{L}$ denotes the learnable layer-weight logits, and $\alpha_l$ is the normalized importance weight assigned to the $l$-th encoder layer.

The pooled representation $\bm{g}(\mathcal M) $ is then mapped to the parameters of a diagonal Gaussian posterior:
\begin{equation}
    \boldsymbol{\mu} = f_\mu(\bm{g}(\mathcal M) ), \qquad
    \log \boldsymbol{\sigma}^2 = \mathrm{clamp}\!\left(f_{\log \sigma^2}(\bm{g}(\mathcal M) ),\, v_{\min},\, v_{\max}\right),
\end{equation}
where $f_\mu$ and $f_{\log \sigma^2}$ are two learnable projection heads that predict the mean and log-variance, respectively. The clamp operation constrains the predicted log-variance within $[v_{\min}, v_{\max}]$ to improve numerical stability. Finally, we sample the latent code using the reparameterization trick~\citep{kingma2014auto}:
\begin{equation}
     \bm{z} = \boldsymbol{\mu} + \boldsymbol{\sigma} \odot \boldsymbol{\epsilon},
    \qquad
    \boldsymbol{\epsilon} \sim \mathcal{N}(\bm{0}, \bm{I}),
\end{equation}
where $\odot$ denotes element-wise multiplication, $\boldsymbol{\sigma}$ is obtained from the predicted log-variance, and $\bm{z}$ is the stochastic latent representation used for generation.
The representation head is optimized using two loss sources: 1) Generator Loss: During representation-conditioned generator training, the generator attempts to predict the clean molecule from its noisy state~\citep{hoogeboom2022equivariant, li2024geometric}, conditioned on the specific molecule’s representation $\bm{z}$ extracted above. For this purpose, the representation head must learn to extract the most meaningful molecule representation for generative tasks, thereby enabling targeted refinement of the final representation and providing a better representation space. 2) KL-Loss $\mathcal{L}_{\rm KL}$~\citep{kingma2014auto}, which enforces $\mu$ and $\sigma$ to follow a normal distribution, leading to a more compact representation space. This process resembles the training of a Variational Autoencoder (VAE)~\citep{kingma2014auto}, where the latent space, in our case, corresponds to the molecule representation itself, and the decoded output is the 3D molecular structure.

\textbf{Molecule Perceptual Loss.} \quad Inspired by perceptual losses in image generation~\citep{zhang2018unreasonable} and the semantic information contained in the shallow layers of the encoder, as discussed in~\Cref{sec:motivation}, we introduce a high-level perceptual representation loss $\mathcal{L}_{\textsc{perc}}$. Instead of solely minimizing the low-level mean squared loss between the denoised and clean molecule coordinates, this loss aligns the representation of the denoised molecule with that of the clean molecule.
Specifically, during training, both the denoised molecule $\hat{\mathcal{M}}$ and the clean molecule $\mathcal{M}$ are passed through the frozen pretrained encoder to extract graph-level representations. The perceptual loss is defined as
\begin{equation}
    \mathcal{L}_{\textsc{perc}} =
    w(t)\left\|
    \mathrm{sg}\!\left(\bm g(\mathcal{M})\right)
    -
    \bm g(\hat{\mathcal{M}})
    \right\|_2^2,
\end{equation}
where $\bm g(\cdot)$ denotes the pooled graph-level representation extracted by the frozen encoder, $\mathrm{sg}(\cdot)$ denotes the stop-gradient operation, and $w(t)$ is a time-dependent weighting function. The stop-gradient operation treats the clean-molecule representation $\bm g(\mathcal{M})$ as a fixed semantic target, while allowing gradients to flow through $\bm g(\hat{\mathcal{M}})$ to optimize the generator. Although the encoder parameters are frozen, gradients can still be backpropagated through the encoder computation graph to the generated molecular coordinates and atom-type logits. Since $\hat{\mathcal{M}}$ also contains discrete atom types, which cannot be directly optimized by standard backpropagation, we employ a straight-through estimator~\citep{bengio2013estimating} to approximate gradients through this discrete path. Finally, $w(t)$ is implemented as a cosine schedule that assigns larger weights to later timesteps, where the generated structure is closer to clean data and the encoder representations provide a more informative alignment signal.

\textbf{Representation Alignment between Generator and Encoder.} \quad To further leverage the encoder beyond graph-level conditioning, motivated by the preceding analysis in~\Cref{sec:motivation}, we introduce a representation projection alignment loss ($\mathcal{L}_{\textsc{repa}}$) to align generator features with pretrained encoder representations at the node level. Let $\bm{H}_{\mathrm{enc}}^{(l_e)} \in \mathbb{R}^{N \times d_e}$ and $\bm{H}_{\mathrm{gen}}^{(l_g)} \in \mathbb{R}^{N \times d_g}$ denote node features from the $l_e$-th encoder layer and $l_g$-th generator layer, respectively. To reconcile the differences in feature dimensions and scales, we employ a learnable projection $\textsc{Proj}: \mathbb{R}^{d_g} \to \mathbb{R}^{d_e}$ to map generator features into the encoder space, and maximize their cosine similarity over valid atoms $\mathcal{A}$:
\begin{equation}
    \mathcal{L}_{\textsc{repa}} = - \frac{1}{|\mathcal{A}|} \sum_{i \in \mathcal{A}} \left\langle \frac{\textsc{Proj}({\bm{H}}_{\mathrm{gen},i}^{(l_g)})}{\left\|\textsc{Proj}({\bm{H}}_{\mathrm{gen},i}^{(l_g)})\right\|_2}, \frac{{\mathrm{sg}(\bm{H}}_{\mathrm{enc},i}^{(l_e)})}{\left\|{\mathrm{sg}(\bm{H}}_{\mathrm{enc},i}^{(l_e)})\right\|_2} \right\rangle,
\end{equation}
Compared to MSE, cosine similarity is adopted to mitigate the influence of feature-scale discrepancies across different representation spaces \citep{yu2024repa}.

Combining all the previous designs leads to the final full Phase-I training objective:
\begin{align}
    \mathcal{L}_{\text{total}}
    &=\mathcal{L}_{\text{gen}}+ \lambda_{\text{KL}} \mathcal{L}_{\text{KL}} + \lambda_{\textsc{perc}} \,{\mathcal{L}}_{\textsc{perc}}
    + \lambda_{\textsc{repa}} \,\mathcal{L}_{\textsc{repa}}.
\end{align}
Here, $\mathcal{L}_{\text{gen}}$ depends on the chosen generative backbone. The complete training algorithm and additional implementation details are provided in \Cref{sec:method-details}.

\textbf{Inference.}\quad At inference time, generation follows GeoRCG~\citep{li2024geometric}, which proceeds in two steps. First, the RDM samples a latent representation $\hat{\boldsymbol{\mu}} \sim p_\phi$, optionally conditioned on the desired atom count, molecular properties or other constraints. Second, the molecule generator produces a 3D structure $\hat{\mathcal{M}} \sim p_\theta(\cdot \mid \hat{\boldsymbol{\mu}})$ conditioned on the sampled representation. This decoupled design allows latent generation and structure generation to be modeled separately.
\label{sec:inference}

\section{Experiments}
\label{sec:experiments}

We evaluate \modelname{} on two standard benchmarks for 3D molecule generation: GEOM-DRUG~\citep{axelrod2022geom} and QM9~\citep{ramakrishnan2014quantum}. Beyond generation quality, we conduct \hlblue{representation-quality analyses}, including effective rank and Lipschitz constant estimation and downstream regression, to validate our core claim that improving representation quality is key to improving generation quality. We further demonstrate that the refined representations transfer to a broader range of tasks, such as regression, thereby pointing to a potential paradigm for \hlorange{generative pretraining}.

\subsection{Experimental Setup}
\label{sec:exp_setup}

\textbf{Datasets.} \quad 
\textbf{QM9}~\citep{ramakrishnan2014quantum} is a benchmark of 133,885 small organic molecules with up to nine heavy atoms, each associated with a DFT-optimized equilibrium geometry and quantum chemical properties computed at the B3LYP/6-31G(2df,p) level. Its high-quality geometries and relatively simple chemical space make it a standard testbed for evaluating 3D molecular generation on small molecules.
\textbf{GEOM-DRUG}~\citep{axelrod2022geom} is a large-scale 3D molecular conformation benchmark of flexible, drug-like molecules, featuring larger molecular size, richer stereochemistry, and more complex conformational landscapes than QM9. It therefore provides a realistic and challenging testbed for evaluating 3D molecular generation on drug-like compounds. 

\textbf{Backbone and Encoder.} \quad 
We take GeoRCG~\citep{li2024geometric}, built upon the SemlaFlow~\citep{irwin2024semlaflow} framework, as our base generation model. We instantiate the pretrained encoder with either UniMol~\citep{unimol} or Frad~\citep{frad}.
The RDM is trained in the representation-head's $\boldsymbol{\mu}$-space with a SimpleMLP backbone.

\textbf{Baselines.} \quad
We compare our method with recent state-of-the-art 3D molecular generation methods: FlowMol~\citep{flowmol}, MiDi~\citep{midi}, JODO~\citep{jodo}, EQGAT-diff~\citep{eqgatdiff}, SemlaFlow~\citep{irwin2024semlaflow}, CanonFlow~\citep{zhou2026rethinking}, and GeoRCG~\citep{li2024geometric}. Baseline details are provided in Appendix~\ref{appendix:baselines}.

\subsection{Main Results}
Table~\ref{tab:drug} presents the results on the GEOM-DRUG.
\modelname{} achieves \emph{state-of-the-art} performance across all reported metrics, surpassing all baselines.
Notably, \modelname{}  achieves 97.28\% validity (vs. the previous best GeoRCG (95.3\%) and CanonFlow (95.9\%)) and 98.51\% molecule stability, while simultaneously achieving the lowest energy and strain energy.
The improvement over the original GeoRCG (from 95.3\% to 97.28\% validity, and from 97.20\% to 98.51\% molecule stability) directly demonstrates the benefit of improved representation quality.
Energy and strain metrics, which reflect the physical plausibility of generated 3D conformations, also improve consistently, confirming that our method leads to more chemically realistic structures.

\begin{table}[t]
\centering
    \caption{Performance of \modelname{} on GEOM-DRUG dataset. Atom stability, molecule stability, validity, and uniqueness are reported as percentages; Energy and strain are reported in kcal/mol. }
\label{tab:drug}
\resizebox{0.93\textwidth}{!}{%
\begin{tabular}{@{}lccccccc@{}}
\toprule
Method & Atom Stab $\uparrow$ & Mol Stab $\uparrow$ & Valid $\uparrow$ & Unique $\uparrow$ &  Energy $\downarrow$ & Strain $\downarrow$ \\
\midrule
FlowMol & 99.0 & 67.5 & 51.2 & ---  & --- & --- \\
MiDi & 99.8 & 91.6 & 77.8 & \textbf{100.0}  & --- & --- \\
EQGAT-diff & \pmerr{99.8}{0.0} & \pmerr{93.4}{0.21} & \pmerr{94.6}{0.24} & \textbf{\pmerr{100.0}{0.0}}  & \pmerr{148.8}{0.9} & \pmerr{140.2}{0.7} \\
SemlaFlow  & \pmerr{99.8}{0.0} & \pmerr{97.3}{0.08} & \pmerr{93.9}{0.19} & \textbf{\pmerr{100.0}{0.0}}  & --- & --- \\
CanonFlow &  \pmerr{99.8}{0.00} & \pmerr{98.4}{0.02} & \pmerr{95.9}{0.08} &  \textbf{\pmerr{100.0}{0.00}} & --- & --- \\
GeoRCG & \pmerr{99.8}{0.00} & \pmerr{97.20}{0.00} & \pmerr{95.3}{0.13} &  \textbf{\pmerr{100.0}{0.00}} & \pmerr{88.6}{1.03} & \pmerr{47.64}{1.10} \\
\midrule
\rowcolor{bgpurple!15}\textbf{\modelname{}  (Ours)} & \textbf{\pmerr{99.88}{0.01}} & \textbf{\pmerr{98.51}{0.16}} & \textbf{\pmerr{97.28}{0.18}} & \textbf{\pmerr{100.00}{0.00}} & \textbf{\pmerr{85.53}{0.45}} & \textbf{\pmerr{43.86}{0.27}} \\
\bottomrule
\end{tabular}
\vspace{-1em}
}
\end{table}

\begin{table*}[t]
    \centering
    \vspace{-3mm}
    \caption{Performance of \modelname{} on QM9 dataset. Atom stability, molecule stability, and validity are reported as percentages; Opt-RMSD is reported in \AA. 
    }
    \small
    \label{tab:unconditional_qm9}
    \vspace{0.2em}
    \resizebox{0.77\textwidth}{!}{
    \begin{tabular}{l|cccc|c}
    \toprule
    Methods              & Atom Stab $\uparrow$        & Mol Stab $\uparrow$            & Valid $\uparrow$        & Opt-RMSD $\downarrow$ & NFE $\downarrow$     \\
    \midrule
    
    EDM   & 98.7 & 82.0 & 91.9& -- & 1000   \\
    GCDM & 98.7 & 85.7 & 94.8 & -- & 1000\\
    MUDiff & 98.3 & 89.9 & 95.3 & -- & 1000 \\
    JODO & 99.9 & 98.8 & 96.0 & -- & 1000\\
    \midrule
    
    FlowMol &  99.7 & 96.2 & 97.3 & -- & 100\\
    MiDi               & 99.8                  & 97.5                & 97.9              & --              & 500        \\
    EQGAT-diff         & \pmerr{99.9}{0.0}           & \pmerr{98.7}{0.18}           & \pmerr{99.0}{0.16}       & -- & 500      \\
    Semlaflow & \pmerr{99.9}{0.0} & \pmerr{99.56}{0.07} & \pmerr{99.31}{0.08} &  \pmerr{0.23}{0.0} & 100\\

    \midrule
    \rowcolor{bgpurple!15}\textbf{\modelname{}  (Ours)} & {\textbf{\pmerr{99.9}{0.0}}}          & \textbf{\pmerr{99.68}{0.04}}           &  \textbf{\pmerr{99.48}{0.08}}     & \textbf{\pmerr{0.21}{0.0}} & 100  \\ 
    \bottomrule
    \end{tabular}
    }
    \vspace{-10pt}
\end{table*}

Table~\ref{tab:unconditional_qm9} reports the results on QM9. \modelname{} achieves strong performance across all four metrics while matching the fewest NFE among compared methods. This result supports our hypothesis that a better-structured latent space can provide more informative conditioning signals, thereby enabling the generator to achieve stronger generalization.

\begin{wrapfigure}[27]{R}{0.55\textwidth}
    \centering
    \vspace{-5pt}
    \includegraphics[width=0.55\textwidth]{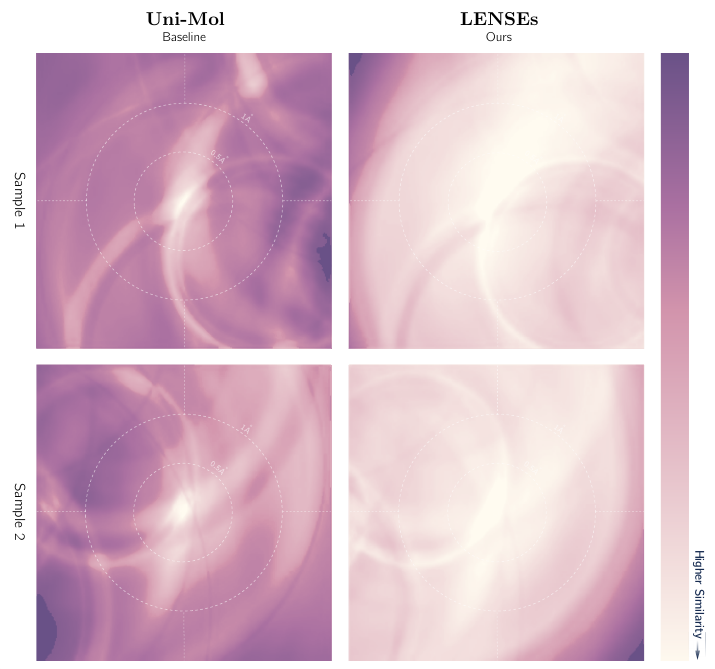}
    \vspace{-7pt}
    \caption{Latent space visualization. The two rows show local latent-space visualizations around two randomly selected atoms from two molecules. The color encodes $1-\frac{\langle \bm h, \bm h_0\rangle}{\|\bm h\|\|\bm h_0\|}$, where $\bm h_0$ is the representation of the center point. Lighter purple regions indicate smaller values, i.e., higher similarity to the center point, reflecting a smoother local representation geometry.}
    \label{fig:latent}
\end{wrapfigure}

\subsection{Representation Quality Analysis}
\label{sec:exp_rep_quality}

We provide two complementary pieces of evidence that \modelname{} yields a
better-conditioned representation space: (1) latent-geometry measurements
showing a less degenerate and more locally stable manifold; (2) downstream
regression performance demonstrating richer representations.

\paragraph{Latent Geometry of the Representation Space.}
\label{sec:exp_lipschitz}

We first measure the \emph{effective rank} of the representation, which
counts the number of effectively independent dimensions in the latent space.
As shown in \Cref{tab:lipschitz}, the effective rank increases from 65.32 to
88.42 on GEOM-DRUG ({1.35$\times$}), and from 95.49 to 109.40 on QM9. This
indicates that the raw encoder representation occupies only a narrow
subspace, while the representation head expands it into a richer,
better-separated manifold.

A second, complementary property is local stability. The Lipschitz constant
of a mapping measures how much the representation responds to input
perturbations: a lower value indicates that small perturbations produce
proportionally small latent changes, which is desirable for RDM sampling
stability. Estimating local Lipschitz constants following the procedure
described in \Cref{alg:empirical-lipschitz}, we find that the latent space
learned by \modelname{} is substantially more stable than the raw encoder
representation, with the mean estimate reduced by $4.6\times$ on GEOM-DRUG
and by $8.7\times$ on QM9 (\Cref{tab:lipschitz}).

For intuition, we further visualize the latent space in \autoref{fig:latent},
where the flattened region around stable conformations is substantially
larger than that of the baseline.

\begin{table}[h]
  \centering
  \vspace{-0.73em}
  \caption{Representation quality on QM9 and GEOM-DRUG.}
  \label{tab:lipschitz}
  \vspace{-0.58em}
  \resizebox{0.77\textwidth}{!}{
  \begin{tabular}{@{}lcc>{\columncolor{lightbluehl}}ccc>{\columncolor{lightbluehl}}c@{}}
    \toprule
    \multirow{2}{*}{Representation}
    & \multicolumn{3}{c}{QM9 \emph{\small with Frad}}
    & \multicolumn{3}{c}{GEOM-DRUG \emph{\small with UniMol}} \\
    \cmidrule(lr){2-4}
    \cmidrule(lr){5-7}
    & \small Lip. mean
    & \small  Lip. max
    &\small  Eff. rank
    &\small  Lip. mean
    &\small  Lip. max
    &\small  Eff. rank \\
    \midrule
    Raw Encoder
    & 249.31
    & 4600.07
    & 95.49
    & 2659.81
    & 3545.67
    & 65.32 \\
    \textbf{\modelname{}}
    & {28.77}
    & {452.25}
    & \textbf{109.40}
    & {570.49}
    & {1781.41}
    & \textbf{88.42} \\
    \bottomrule
  \end{tabular}}
  \vspace{-1.25em}
\end{table}

\paragraph{Downstream Regression on QM9.}
\label{sec:exp_regression}

To examine whether the learned latent representations remain useful beyond generation, we freeze the pretrained encoder and the representation head of \modelname{}, and train only a lightweight readout on QM9 property prediction tasks.

As shown in \Cref{tab:frozen-transfer}(b), \modelname{} consistently improves frozen-representation regression over the raw Frad encoder across the evaluated QM9 properties.
This indicates that representation-aligned \hlorange{generative training} does not merely benefit sampling, but also refines the encoder features into a more predictive molecular representation.

\subsection{Frozen Representation Transfer Across Datasets}
\label{sec:frozen-transfer}

The Frad results above show that the learned representation supports property prediction even when the encoder and representation head are frozen. Here we test whether this benefit carries over to other datasets. In every evaluation below, the pretrained encoder, the learned pooling, and the representation head stay frozen, and we fit only a task-specific linear probe. These experiments therefore measure how useful a representation adapter learned through generative training is downstream, without updating the encoder or the head on downstream labels.

\paragraph{QM9 property regression.}
\Cref{tab:frozen-transfer}(a) reports frozen-probing results on 12 QM9 properties. The learned representation lowers MAE on every property, with relative reductions from 21.6\% to 92.4\%. Because this evaluation reuses the head trained on GEOM-DRUG, it is a cross-dataset transfer setting that complements the original Frad experiment in \Cref{tab:frozen-transfer}(b).

\paragraph{MoleculeNet classification and regression.}
We also evaluate the frozen UniMol-based adapter on nine MoleculeNet datasets~\citep{wu2017moleculenet}: six classification datasets scored by ROC-AUC and three regression datasets scored by RMSE (\Cref{tab:frozen-transfer}(c)). The refined representation improves results on every dataset. Mean ROC-AUC across the classification datasets rises from 75.91\% to 80.18\%, and RMSE falls by 32.8\% on ESOL, 26.5\% on FreeSolv, and 24.5\% on Lipophilicity. For multi-task datasets, each reported score is the average over that dataset's endpoints.

\begin{table*}[t]
\centering
\caption{
\textbf{Frozen-transfer performance.}
Green values indicate relative regression-error reductions (\%)
or absolute ROC-AUC gains (pp).
QM9 energy errors are reported in meV.
}
\vspace{0.5em}
\label{tab:frozen-transfer}

\footnotesize
\setlength{\tabcolsep}{2.5pt}

\resizebox{\textwidth}{!}{%
\setlength{\tabcolsep}{6pt}
\begin{tabular}{@{\extracolsep{\fill}}c*{12}{c}@{}}
\toprule
\multicolumn{13}{@{}l}{%
    \textbf{(a) UniMol $\rightarrow$ QM9}
    \quad \emph{MAE $\downarrow$}
} \\
\midrule
\multirow[c]{2}{*}{Method}
& $\mu$ & $\alpha$ & $\Delta\varepsilon$ & $\varepsilon_{\mathrm{HOMO}}$ & $\varepsilon_{\mathrm{LUMO}}$ & $\langle R^2\rangle$ & ZPVE & $C_v$ & $U_0$ & $U$ & $H$ & $G$ \\
& {\scriptsize D} & {\scriptsize bohr$^3$} & {\scriptsize meV} & {\scriptsize meV} & {\scriptsize meV} & {\scriptsize bohr$^2$} & {\scriptsize meV} & {\scriptsize cal/mol/K} & {\scriptsize meV} & {\scriptsize meV} & {\scriptsize meV} & {\scriptsize meV} \\
\midrule
Baseline
& 0.8096 & 2.0130 & 587.7 & 281.8 & 503.1 & 61.0551
& 85.0 & 1.0032 & 2.2913 & 2.1217 & 2.2519 & 2.1076 \\
\rowcolor{bgpurple!15}
\multicolumn{1}{@{}>{\columncolor{bgpurple!15}[0pt][\tabcolsep]}c}{} & \textbf{0.6344} & \textbf{0.5584} & \textbf{330.3} & \textbf{190.3} & \textbf{265.6} & \textbf{11.1059} & \textbf{6.50} & \textbf{0.1719} & \textbf{0.2205} & \textbf{0.2248} & \textbf{0.2230} & \multicolumn{1}{>{\columncolor{bgpurple!15}[\tabcolsep][0pt]}c@{}}{\textbf{0.2228}} \\
\rowcolor{bgpurple!15}
\multicolumn{1}{@{}>{\columncolor{bgpurple!15}[0pt][\tabcolsep]}c}{\multirow[c]{-2}{*}{LENSEs}} & {\scriptsize\textcolor{gainGreen}{21.6\%$\uparrow$}} & {\scriptsize\textcolor{gainGreen}{72.3\%$\uparrow$}} & {\scriptsize\textcolor{gainGreen}{43.8\%$\uparrow$}} & {\scriptsize\textcolor{gainGreen}{32.5\%$\uparrow$}} & {\scriptsize\textcolor{gainGreen}{47.2\%$\uparrow$}} & {\scriptsize\textcolor{gainGreen}{81.8\%$\uparrow$}} & {\scriptsize\textcolor{gainGreen}{92.4\%$\uparrow$}} & {\scriptsize\textcolor{gainGreen}{82.9\%$\uparrow$}} & {\scriptsize\textcolor{gainGreen}{90.4\%$\uparrow$}} & {\scriptsize\textcolor{gainGreen}{89.4\%$\uparrow$}} & {\scriptsize\textcolor{gainGreen}{90.1\%$\uparrow$}} & \multicolumn{1}{>{\columncolor{bgpurple!15}[\tabcolsep][0pt]}c@{}}{{\scriptsize\textcolor{gainGreen}{89.4\%$\uparrow$}}} \\
\bottomrule
\end{tabular}%
}
\par

\vspace{0.25em}
\resizebox{\textwidth}{!}{
    \setlength{\tabcolsep}{14pt}
\begin{tabular}{@{\extracolsep{\fill}}c*{8}{c}@{}}
\multicolumn{9}{@{}l}{
\textbf{(b) Frad $\rightarrow$ QM9}
\quad \emph{MAE $\downarrow$}
} \\
\midrule
\multirow[c]{2}{*}{Method}
& $\mu$ & $\alpha$ & $\Delta\varepsilon$ & $\varepsilon_{\mathrm{HOMO}}$ & $\varepsilon_{\mathrm{LUMO}}$ & $\langle R^2\rangle$ & ZPVE & $C_v$ \\
& {\scriptsize D} & {\scriptsize bohr$^3$} & {\scriptsize meV} & {\scriptsize meV} & {\scriptsize meV} & {\scriptsize bohr$^2$} & {\scriptsize meV} & {\scriptsize cal/mol/K} \\
\midrule
Baseline
& 0.152 & 0.247 & 96.67 & 58.70
& 54.65 & 7.23 & 11.58 & 0.100 \\
\rowcolor{bgpurple!15}
\multicolumn{1}{@{}>{\columncolor{bgpurple!15}[0pt][\tabcolsep]}c}{} & \textbf{0.119} & \textbf{0.134} & \textbf{68.76} & \textbf{47.87} & \textbf{41.90} & \textbf{3.33} & \textbf{3.69} & \multicolumn{1}{>{\columncolor{bgpurple!15}[\tabcolsep][0pt]}c@{}}{\textbf{0.0481}} \\
\rowcolor{bgpurple!15}
\multicolumn{1}{@{}>{\columncolor{bgpurple!15}[0pt][\tabcolsep]}c}{\multirow[c]{-2}{*}{LENSEs}} & {\scriptsize\textcolor{gainGreen}{21.7\%$\uparrow$}} & {\scriptsize\textcolor{gainGreen}{45.7\%$\uparrow$}} & {\scriptsize\textcolor{gainGreen}{28.9\%$\uparrow$}} & {\scriptsize\textcolor{gainGreen}{18.4\%$\uparrow$}} & {\scriptsize\textcolor{gainGreen}{23.3\%$\uparrow$}} & {\scriptsize\textcolor{gainGreen}{53.9\%$\uparrow$}} & {\scriptsize\textcolor{gainGreen}{68.1\%$\uparrow$}} & \multicolumn{1}{>{\columncolor{bgpurple!15}[\tabcolsep][0pt]}c@{}}{{\scriptsize\textcolor{gainGreen}{51.9\%$\uparrow$}}} \\
\bottomrule
\end{tabular}
}
\par

\vspace{0.15em}
\resizebox{\textwidth}{!}{
        \setlength{\tabcolsep}{8pt}
\begin{tabular}{
@{\extracolsep{\fill}}c*{7}{c}|*{3}{c}@{}
}
\multicolumn{11}{@{}l}{
\textbf{(c) UniMol $\rightarrow$ MoleculeNet}
} \\
\midrule
\multirow[c]{2}{*}[-2pt]{Method}
& \multicolumn{7}{c}{
\emph{Classification: ROC-AUC (\%) $\uparrow$}
}
& \multicolumn{3}{c}{
\emph{Regression: RMSE $\downarrow$}
} \\
\cmidrule(lr){2-8}
\cmidrule(l){9-11}

& BACE & BBBP & ClinTox & HIV
& Tox21 & SIDER & \textbf{Average}
& ESOL & FreeSolv & Lipophilicity \\
\midrule
Baseline
& 75.44 & 89.13 & 78.17 & 74.51
& 79.33 & 58.85 & 75.91
& 0.836 & 1.867 & 0.870 \\
\rowcolor{bgpurple!15}
\multicolumn{1}{@{}>{\columncolor{bgpurple!15}[0pt][\tabcolsep]}c}{} & \textbf{82.02} & \textbf{89.43} & \textbf{85.02} & \textbf{75.90} & \textbf{81.99} & \textbf{66.70} & \textbf{80.18} & \textbf{0.562} & \textbf{1.372} & \multicolumn{1}{>{\columncolor{bgpurple!15}[\tabcolsep][0pt]}c@{}}{\textbf{0.657}} \\
\rowcolor{bgpurple!15}
\multicolumn{1}{@{}>{\columncolor{bgpurple!15}[0pt][\tabcolsep]}c}{\multirow[c]{-2}{*}{LENSEs}} & {\scriptsize\textcolor{gainGreen}{6.58\,pp$\uparrow$}} & {\scriptsize\textcolor{gainGreen}{0.30\,pp$\uparrow$}} & {\scriptsize\textcolor{gainGreen}{6.85\,pp$\uparrow$}} & {\scriptsize\textcolor{gainGreen}{1.39\,pp$\uparrow$}} & {\scriptsize\textcolor{gainGreen}{2.66\,pp$\uparrow$}} & {\scriptsize\textcolor{gainGreen}{7.85\,pp$\uparrow$}} & {\scriptsize\textcolor{gainGreen}{4.27\,pp$\uparrow$}} & {\scriptsize\textcolor{gainGreen}{32.8\%$\uparrow$}} & {\scriptsize\textcolor{gainGreen}{26.5\%$\uparrow$}} & \multicolumn{1}{>{\columncolor{bgpurple!15}[\tabcolsep][0pt]}c@{}}{{\scriptsize\textcolor{gainGreen}{24.5\%$\uparrow$}}} \\
\bottomrule
\end{tabular}
}
\end{table*}

Combined with the Frad results in \Cref{tab:frozen-transfer}(b), the refined representation outperforms the raw encoder features under both encoders and in all 21 frozen-transfer comparisons. These gains therefore indicate that generative training with alignment objectives learns a transferable representation adapter on top of a frozen encoder, rather than features specialized to the generation dataset.

\subsection{Ablation Study}
\label{ablation}

\begin{table}[t]
    \centering
    \caption{
    Ablation of representation aggregation and loss weights on GEOM-DRUG. Loss-weight rows use learned pooling unless marked as final-layer variants. The REPA block reports a local sweep; the original logarithmic sweep and full metrics are given in \Cref{tab:full-ablation-grid}.
    }
    \label{tab:main-ablation}
    \resizebox{\linewidth}{!}{
    \begin{tabular}{llccc>{\columncolor{lightbluehl}}ccccc}
    \toprule
    Part
    & Setting
    & $\lambda_{\mathrm{KL}}$
    & $\lambda_{\mathrm{perc}}$
    & $\lambda_{\mathrm{REPA}}$
    & Validity (\%) $\uparrow$
    & Mol. Stability (\%) $\uparrow$
    & Energy $\downarrow$
    & Opt. RMSD $\downarrow$
    & Strain $\downarrow$ \\
    \midrule

    \makecell[l]{\textbf{Baseline} {\scriptsize(reproduced)}}
    & Final layer
    & -- & -- & --
    & 94.20 & 97.20 & 103.33 & 0.8730 & 66.95 \\
        \textbf{Full \modelname{}}
    & Learned pooling
    & $5\times 10^{-7}$
    & $1\times 10^{-3}$
    & $1\times10^{-3}$
    & \textbf{97.28}
    & 98.51
    & 85.53
    & 0.7766
    & 43.86 \\
    \midrule

    \multirow{5}{*}{\makecell[l]{\textbf{KL loss}\\\textit{w/o Perc. and REPA}}}
    & KL
    & $1\times10^{-4}$ & -- & --
    & 0.00 & 0.00 & NaN & NaN & NaN \\

    & KL
    & $1\times10^{-5}$ & -- & --
    & 93.40 & 96.35 & 110.58 & 0.9301 & 70.28 \\

    & KL
    & $1\times10^{-6}$ & -- & --
    & 95.00 & 96.63 & 107.43 & 0.9395 & 64.60 \\

    & KL
    & $5\times10^{-7}$ & -- & --
    & 96.52 & 98.01 & 90.66 & 0.7780 & 43.38 \\

    & KL
    & $1\times10^{-8}$ & -- & --
    & 95.53 & 97.22 & 93.04 & 0.8564 & 47.00 \\

    \midrule

    \multirow{5}{*}{\makecell[l]{\textbf{Perc. loss}\\\textit{w/o REPA}}}
    & KL + Perc.
    & $5\times10^{-7}$ & $5\times10^{-4}$ & --
    & 94.20 & 98.60 & 79.29 & 0.8303 & 45.00 \\

    & KL + Perc.
    & $5\times10^{-7}$ & $7.5\times10^{-4}$ & --
    & 96.00 & 98.30 & 80.18 & 0.8455 & 42.55 \\

    & KL + Perc.
    & $5\times10^{-7}$ & $1\times10^{-3}$ & --
    & 96.60 & 98.70 & 82.63 & 0.8317 & 39.98 \\

    & KL + Perc.
    & $5\times10^{-7}$ & $5\times10^{-3}$ & --
    & 93.80 & 98.00 & 87.93 & 0.9572 & 40.79 \\

    & KL + Perc.
    & $5\times10^{-7}$ & $1\times10^{-2}$ & --
    & 96.50 & 98.80 & 91.17 & 0.8303 & 44.79 \\

    \midrule

    \multirow{4}{*}{\makecell[l]{\textbf{REPA loss}\\\textit{w/o Perc.}}}
    & KL + REPA
    & $5\times10^{-7}$ & -- & $5\times10^{-4}$
    & 96.60 & 98.40 & 87.34 & 0.8375 & 45.01 \\

    & KL + REPA
    & $5\times10^{-7}$ & -- & $1\times10^{-3}$
    & 96.40 & 98.00 & 86.87 & 0.8091 & 45.83 \\

    & KL + REPA
    & $5\times10^{-7}$ & -- & $1.5\times10^{-3}$
    & 96.20 & 98.70 & 88.12 & 0.8412 & 46.92 \\

    & KL + REPA
    & $5\times10^{-7}$ & -- & $2\times10^{-3}$
    & 96.10 & 97.70 & 87.02 & 0.8253 & 47.73 \\
\midrule
    {w/o Multi-layer Agg.}
    & KL
    & $5\times 10^{-7}$ & -- & --
    & 95.90 & 97.80 & 93.25 & 0.8138 & 46.29 \\

    {w/o Multi-layer Agg.}
    & All
    & $5\times 10^{-7}$
    & $1\times 10^{-3}$
    & $1\times10^{-3}$
    & 97.10 & 98.20 & 86.20 & 0.8200 & 44.72 \\
    \bottomrule
    \end{tabular}
    }
    \vspace{-1em}
\end{table}

We ablate the three key objectives in \modelname{}: the KL loss, the perceptual loss, and the REPA loss.
For the KL loss, too large a weight can collapse the representation, while too small a weight under-regularizes the latent space.
For the perceptual and REPA losses, large weights can overwhelm the generation objective or introduce conflicting gradients, while small weights weaken encoder-space semantic supervision and node-level alignment, respectively.
\modelname{} is robust to its loss weights over wide ranges. Every REPA weight across four orders of magnitude ($10^{-6}$--$10^{-2}$) and every perceptual weight across a $20\times$ range ($5\times10^{-4}$--$10^{-2}$) lowers both energy and strain relative to the baseline, and all REPA settings also improve validity (\Cref{tab:main-ablation,tab:full-ablation-grid}). The KL weight is likewise stable across nearly two orders of magnitude ($10^{-8}$--$5\times10^{-7}$); only beyond this range does the penalty pull the posterior toward the prior and degrade the conditioning signal, as is typical for KL regularization. \Cref{tab:main-ablation} also compares final-layer aggregation with our learned multi-layer pooling.
All reproduced variants share the same data, training budget, codebase, checkpoint-selection rule, and evaluation protocol.

Taken together, the ablations show that the components contribute complementary benefits.
The KL-regularized representation head smooths the conditioning latent space, improving validity and physical plausibility.
The perceptual loss reduces energy and strain, and REPA adds node-level alignment that further improves the overall trade-off when combined with the other components.
The full \modelname{} achieves the \emph{highest validity}, maintains strong stability, and has substantially lower energy and strain than the baseline.

\Cref{sec:appendix-results} gives further analyses. Beyond, \Cref{app:reduced_generation_steps} shows that the refined representations keep generation quality strong with only a few sampling steps.

\section{Conclusion}
\label{sec:conclusion}
We presented \modelname{}, a framework for geometric representation-conditioned molecule generation that improves generation by explicitly refining the conditioning latent space. Our work is motivated by the observation that useful molecular representations are not necessarily generation-friendly: pretrained encoder semantics may remain underutilized, poorly aligned with generator dynamics, or embedded in a latent space with suboptimal geometry.

\modelname{} addresses this gap through three complementary components. The representation head with learnable layer pooling extracts smoother and more expressive multi-layer encoder features. The perceptual-style representation loss transfers multi-scale molecular semantics by aligning generated and ground-truth molecules in the pretrained encoder feature space. The REPA loss further aligns the generator's hidden states with encoder representations at the node level.

Experiments on GEOM-DRUG demonstrate that \modelname{} achieves SOTA structural validity and physical plausibility. These gains come with a better-conditioned representation space: effective rank increases, and the mean local Lipschitz estimate falls on GEOM-DRUG and QM9. These results suggest that high-quality generation depends not only on the generator, but also on a conditioning representation that is smooth, semantically rich, and aligned with the generative objective.

Beyond generation, our results show that \emph{generative training is an effective way to pretrain molecular representations}. With the encoder and representation head frozen, the refined features outperform the raw encoder features on every QM9 property under both Frad and UniMol, and on all MoleculeNet datasets. These gains include MAE reductions on QM9 and an increase in mean ROC-AUC on MoleculeNet. Because the UniMol head is trained only on GEOM-DRUG and never sees downstream labels, these gains reflect transferable representation learning rather than in-domain fitting. Generation is therefore not only a consumer of pretrained representations but also a source of better ones. We believe future molecular foundation models should treat generation as a core pretraining objective, unifying representation learning, semantic alignment, and generation within a single framework to support both molecular understanding and design.

\textbf{Future Work.} Future work will focus on three directions: (1) applying \modelname{} to more efficient training and inference frameworks, including few-step generation settings~\cite{zhou2026rethinking}; (2) scaling generative pretraining to larger encoders and datasets, and characterizing when and why it improves downstream generalization; and (3) developing molecular-domain perceptual models for more effective semantic extraction.%

\clearpage
\section*{Acknowledgments}
This work is supported by the Beijing Natural Science Foundation (Z260022) and National Natural Science Foundation of China (62550138).

The first author thanks everyone who helped him find his way into AI for Science, and who kept him going when he wasn't sure he had chosen right. To Prof.\ Muhan Zhang and Zian Li, who first brought him into research and patiently guided his early steps; to Qiaojing Huang, for his support and generous help during his internship at ByteDance; and to Yuli She and Kai Liu, for the opportunities and the room to grow.

He is deeply grateful to his parents for their understanding and care through all the late nights. Thanks to Jiangyue Zeng for her companionship, patience, and support throughout, and to all his friends at \emph{Lvzhiqiang} for years of cheering each other on.

To everyone who has been part of this journey: thank you!

\bibliographystyle{unsrtnat}
\bibliography{reference}

\begin{thebibliography}{60}
\providecommand{\natexlab}[1]{#1}
\providecommand{\url}[1]{\texttt{#1}}
\expandafter\ifx\csname urlstyle\endcsname\relax
  \providecommand{\doi}[1]{doi: #1}\else
  \providecommand{\doi}{doi: \begingroup \urlstyle{rm}\Url}\fi

\bibitem[Zeng et~al.(2022)Zeng, Wang, Luo, Kang, Tang, Lightstone, Fang,
  Cornell, Nussinov, and Cheng]{zeng2022deep}
Xiangxiang Zeng, Fei Wang, Yuan Luo, Seung-gu Kang, Jian Tang, Felice~C
  Lightstone, Evandro~F Fang, Wendy Cornell, Ruth Nussinov, and Feixiong Cheng.
\newblock Deep generative molecular design reshapes drug discovery.
\newblock \emph{Cell Reports Medicine}, 3\penalty0 (12), 2022.

\bibitem[Cheng et~al.(2021)Cheng, Gong, Liu, Song, and Zou]{cheng2021molecular}
Yu~Cheng, Yongshun Gong, Yuansheng Liu, Bosheng Song, and Quan Zou.
\newblock Molecular design in drug discovery: a comprehensive review of deep
  generative models.
\newblock \emph{Briefings in bioinformatics}, 22\penalty0 (6):\penalty0
  bbab344, 2021.

\bibitem[Tang et~al.(2024)Tang, Dai, Knight, Wu, Li, Li, and
  Gerstein]{tang2024survey}
Xiangru Tang, Howard Dai, Elizabeth Knight, Fang Wu, Yunyang Li, Tianxiao Li,
  and Mark Gerstein.
\newblock A survey of generative ai for de novo drug design: new frontiers in
  molecule and protein generation.
\newblock \emph{Briefings in Bioinformatics}, 25\penalty0 (4):\penalty0
  bbae338, 2024.

\bibitem[Weininger(1988)]{weininger1988smiles}
David Weininger.
\newblock Smiles, a chemical language and information system. 1. introduction
  to methodology and encoding rules.
\newblock \emph{Journal of Chemical Information and Computer Sciences},
  28\penalty0 (1):\penalty0 31--36, 1988.

\bibitem[Krenn et~al.(2020)Krenn, H{\"a}se, Nigam, Friederich, and
  Aspuru-Guzik]{krenn2020self}
Mario Krenn, Florian H{\"a}se, AkshatKumar Nigam, Pascal Friederich, and Alan
  Aspuru-Guzik.
\newblock Self-referencing embedded strings (selfies): A 100\% robust molecular
  string representation.
\newblock \emph{Machine Learning: Science and Technology}, 1\penalty0
  (4):\penalty0 045024, 2020.

\bibitem[Cheng et~al.(2023)Cheng, Cai, Miret, Malkomes, Phielipp, and
  Aspuru-Guzik]{cheng2023group}
Austin~H Cheng, Andy Cai, Santiago Miret, Gustavo Malkomes, Mariano Phielipp,
  and Al{\'a}n Aspuru-Guzik.
\newblock Group selfies: a robust fragment-based molecular string
  representation.
\newblock \emph{Digital Discovery}, 2\penalty0 (3):\penalty0 748--758, 2023.

\bibitem[Hoogeboom et~al.(2022)Hoogeboom, Satorras, Vignac, and
  Welling]{hoogeboom2022equivariant}
Emiel Hoogeboom, V{\'i}ctor~Garcia Satorras, Clement Vignac, and Max Welling.
\newblock Equivariant diffusion for molecule generation in 3d.
\newblock \emph{International Conference on Machine Learning}, pages
  9087--9102, 2022.

\bibitem[Xu et~al.(2022)Xu, Yu, Song, Shi, Ermon, and Tang]{xu2022geodiff}
Minkai Xu, Lantao Yu, Yang Song, Chence Shi, Stefano Ermon, and Jian Tang.
\newblock Geodiff: A geometric diffusion model for molecular conformation
  generation.
\newblock \emph{arXiv preprint arXiv:2203.02923}, 2022.

\bibitem[Adak et~al.(2025)Adak, Rawat, and Vyas]{adak2025molvision}
Deepan Adak, Yogesh~Singh Rawat, and Shruti Vyas.
\newblock Molvision: Molecular property prediction with vision language models.
\newblock \emph{arXiv preprint arXiv:2507.03283}, 2025.

\bibitem[Wang et~al.(2024)Wang, Zhang, Jiang, Chen, Zhu, Yan, Wu, and
  Xu]{wang2024x2}
Zhanfeng Wang, Wenhao Zhang, Minghong Jiang, Yicheng Chen, Zhenyu Zhu, Wenjie
  Yan, Jianming Wu, and Xin Xu.
\newblock X2-gnn: A physical message passing neural network with natural
  generalization ability to large and complex molecules.
\newblock \emph{The Journal of Physical Chemistry Letters}, 15\penalty0
  (51):\penalty0 12501--12512, 2024.

\bibitem[Ho et~al.(2020)Ho, Jain, and Abbeel]{ho2020denoising}
Jonathan Ho, Ajay Jain, and Pieter Abbeel.
\newblock Denoising diffusion probabilistic models.
\newblock \emph{Advances in Neural Information Processing Systems},
  33:\penalty0 6840--6851, 2020.

\bibitem[Song et~al.(2020)Song, Meng, and Ermon]{song2020denoising}
Jiaming Song, Chenlin Meng, and Stefano Ermon.
\newblock Denoising diffusion implicit models.
\newblock \emph{arXiv preprint arXiv:2010.02502}, 2020.

\bibitem[Lipman et~al.(2022)Lipman, Chen, Ben-Hamu, Nickel, and
  Le]{lipman2022flow}
Yaron Lipman, Ricky~TQ Chen, Heli Ben-Hamu, Maximilian Nickel, and Matt Le.
\newblock Flow matching for generative modeling.
\newblock \emph{arXiv preprint arXiv:2210.02747}, 2022.

\bibitem[Watson et~al.(2023)Watson, Juergens, Bennett, Trippe, Yim, Eisenach,
  Ahern, Borst, Ragotte, Milles, et~al.]{watson2023novo}
Joseph~L Watson, David Juergens, Nathaniel~R Bennett, Brian~L Trippe, Jason
  Yim, Helen~E Eisenach, Woody Ahern, Andrew~J Borst, Robert~J Ragotte, Lukas~F
  Milles, et~al.
\newblock De novo design of protein structure and function with rfdiffusion.
\newblock \emph{Nature}, 620\penalty0 (7976):\penalty0 1089--1100, 2023.

\bibitem[Corso et~al.(2022)Corso, St{\"a}rk, Jing, Barzilay, and
  Jaakkola]{corso2022diffdock}
Gabriele Corso, Hannes St{\"a}rk, Bowen Jing, Regina Barzilay, and Tommi
  Jaakkola.
\newblock Diffdock: Diffusion steps, twists, and turns for molecular docking.
\newblock \emph{arXiv preprint arXiv:2210.01776}, 2022.

\bibitem[Walters and Barzilay(2020)]{walters2020applications}
W~Patrick Walters and Regina Barzilay.
\newblock Applications of deep learning in molecule generation and molecular
  property prediction.
\newblock \emph{Accounts of chemical research}, 54\penalty0 (2):\penalty0
  263--270, 2020.

\bibitem[Tom et~al.(2024)Tom, Schmid, Baird, Cao, Darvish, Hao, Lo,
  Pablo-Garc{\'\i}a, Rajaonson, Skreta, et~al.]{tom2024self}
Gary Tom, Stefan~P Schmid, Sterling~G Baird, Yang Cao, Kourosh Darvish, Han
  Hao, Stanley Lo, Sergio Pablo-Garc{\'\i}a, Ella~M Rajaonson, Marta Skreta,
  et~al.
\newblock Self-driving laboratories for chemistry and materials science.
\newblock \emph{Chemical Reviews}, 124\penalty0 (16):\penalty0 9633--9732,
  2024.

\bibitem[Xue et~al.(2019)Xue, Gong, Yang, Chuai, Qu, Shen, Yu, and
  Liu]{xue2019advances}
Dongyu Xue, Yukang Gong, Zhaoyi Yang, Guohui Chuai, Sheng Qu, Aizong Shen, Jing
  Yu, and Qi~Liu.
\newblock Advances and challenges in deep generative models for de novo
  molecule generation.
\newblock \emph{Wiley Interdisciplinary Reviews: Computational Molecular
  Science}, 9\penalty0 (3):\penalty0 e1395, 2019.

\bibitem[Dunn and Koes(2026)]{dunn2026flowmol3}
Ian Dunn and David~R Koes.
\newblock Flowmol3: flow matching for 3d de novo small-molecule generation.
\newblock \emph{Digital Discovery}, 2026.

\bibitem[Zeng et~al.(2026)Zeng, Jin, Ambrose, Karypis, Transtrum, Tadmor,
  Hennig, Roitberg, Martiniani, and Liu]{zeng2026propmolflow}
Cheng Zeng, Jirui Jin, Connor Ambrose, George Karypis, Mark Transtrum, Ellad~B
  Tadmor, Richard~G Hennig, Adrian Roitberg, Stefano Martiniani, and Mingjie
  Liu.
\newblock Propmolflow: property-guided molecule generation with
  geometry-complete flow matching.
\newblock \emph{Nature Computational Science}, pages 1--10, 2026.

\bibitem[Reidenbach et~al.(2026)Reidenbach, Nikitin, Isayev, and
  Paliwal]{reidenbach2026applications}
Danny Reidenbach, Filipp Nikitin, Olexandr Isayev, and Saee~Gopal Paliwal.
\newblock Applications of modular co-design for de novo 3d molecule generation.
\newblock \emph{Digital Discovery}, 5\penalty0 (2):\penalty0 754--768, 2026.

\bibitem[Poletukhin et~al.(2026)Poletukhin, Kollovieh, Eberhard, and
  G{\"u}nnemann]{poletukhin20263d}
Roman Poletukhin, Marcel Kollovieh, Eike Eberhard, and Stephan G{\"u}nnemann.
\newblock 3d molecule generation from rigid motifs via se (3) flows.
\newblock \emph{arXiv preprint arXiv:2601.16955}, 2026.

\bibitem[Grisoni et~al.(2020)Grisoni, Moret, Lingwood, and
  Schneider]{grisoni2020bidirectional}
Francesca Grisoni, Michael Moret, Robin Lingwood, and Gisbert Schneider.
\newblock Bidirectional molecule generation with recurrent neural networks.
\newblock \emph{Journal of chemical information and modeling}, 60\penalty0
  (3):\penalty0 1175--1183, 2020.

\bibitem[Xu et~al.(2023{\natexlab{a}})Xu, Powers, Dror, Ermon, and
  Leskovec]{xu2023geometric}
Minkai Xu, Alexander~S Powers, Ron~O Dror, Stefano Ermon, and Jure Leskovec.
\newblock Geometric latent diffusion models for 3d molecule generation.
\newblock In \emph{International Conference on Machine Learning}, pages
  38592--38610. PMLR, 2023{\natexlab{a}}.

\bibitem[Li et~al.(2024{\natexlab{a}})Li, Zhou, Wang, Peng, and
  Zhang]{li2024geometric}
Zian Li, Cai Zhou, Xiyuan Wang, Xingang Peng, and Muhan Zhang.
\newblock Geometric representation condition improves equivariant molecule
  generation.
\newblock \emph{arXiv preprint arXiv:2410.03655}, 2024{\natexlab{a}}.

\bibitem[Yang and Wang(2023)]{yang2023diffusion}
Xingyi Yang and Xinchao Wang.
\newblock Diffusion model as representation learner.
\newblock In \emph{Proceedings of the IEEE/CVF International Conference on
  Computer Vision}, pages 18938--18949, 2023.

\bibitem[Zhou et~al.(2023)Zhou, Gao, Ding, Zheng, Xu, Wei, Zhang, and
  Ke]{unimol}
Gengmo Zhou, Zhifeng Gao, Qiankun Ding, Hang Zheng, Hongteng Xu, Zhewei Wei,
  Linfeng Zhang, and Guolin Ke.
\newblock Uni-mol: A universal 3d molecular representation learning framework.
\newblock \emph{International Conference on Learning Representations}, 2023.

\bibitem[Feng et~al.(2023)Feng, Ni, Lan, Ma, and Ma]{frad}
Shikun Feng, Zhi Ni, Yanyan Lan, Zhi-Ming Ma, and Wei-Ying Ma.
\newblock Fractional denoising for 3d molecular pre-training.
\newblock \emph{International Conference on Machine Learning}, 2023.

\bibitem[Li et~al.(2024{\natexlab{b}})Li, Katabi, and He]{li2024return}
Tianhong Li, Dina Katabi, and Kaiming He.
\newblock Return of unconditional generation: A self-supervised representation
  generation method.
\newblock \emph{Advances in Neural Information Processing Systems},
  37:\penalty0 125441--125468, 2024{\natexlab{b}}.

\bibitem[Yan et~al.(2025)Yan, Li, and Zhang]{yan2025georecon}
Shaoheng Yan, Zian Li, and Muhan Zhang.
\newblock Georecon: Graph-level representation learning for 3d molecules via
  reconstruction-based pretraining.
\newblock \emph{arXiv preprint arXiv:2506.13174}, 2025.

\bibitem[Yin et~al.(2025)Yin, Liu, Hao, Zhou, Liu, Ma, and Wang]{yin2025multi}
Rong Yin, Ruyue Liu, Xiaoshuai Hao, Xingrui Zhou, Yong Liu, Can Ma, and Weiping
  Wang.
\newblock Multi-modal molecular representation learning via structure
  awareness.
\newblock \emph{IEEE Transactions on Image Processing}, 2025.

\bibitem[Gatys et~al.(2016)Gatys, Ecker, and Bethge]{gatys2016image}
Leon~A Gatys, Alexander~S Ecker, and Matthias Bethge.
\newblock Image style transfer using convolutional neural networks.
\newblock In \emph{Proceedings of the IEEE conference on computer vision and
  pattern recognition}, pages 2414--2423, 2016.

\bibitem[Lin et~al.(2017)Lin, Doll{\'a}r, Girshick, He, Hariharan, and
  Belongie]{lin2017feature}
Tsung-Yi Lin, Piotr Doll{\'a}r, Ross Girshick, Kaiming He, Bharath Hariharan,
  and Serge Belongie.
\newblock Feature pyramid networks for object detection.
\newblock In \emph{Proceedings of the IEEE conference on computer vision and
  pattern recognition}, pages 2117--2125, 2017.

\bibitem[Johnson et~al.(2016)Johnson, Alahi, and
  Fei-Fei]{johnson2016perceptual}
Justin Johnson, Alexandre Alahi, and Li~Fei-Fei.
\newblock Perceptual losses for real-time style transfer and super-resolution.
\newblock \emph{European Conference on Computer Vision}, pages 694--711, 2016.

\bibitem[Yu et~al.(2024)Yu, Kwak, Jang, Jeong, Huang, Shin, and
  Xie]{yu2024repa}
Sihyun Yu, Sangkyung Kwak, Huiwon Jang, Jongheon Jeong, Jonathan Huang, Jinwoo
  Shin, and Saining Xie.
\newblock Representation alignment for generation: Training diffusion
  transformers is easier than you think.
\newblock \emph{arXiv preprint arXiv:2410.06940}, 2024.

\bibitem[Zhou et~al.(2026)Zhou, Chen, Li, Wang, Jiang, Li, Yu, Zhang, Bates,
  and Jaakkola]{zhou2026rethinking}
Cai Zhou, Zijie Chen, Zian Li, Jike Wang, Kaiyi Jiang, Pan Li, Rose Yu, Muhan
  Zhang, Stephen Bates, and Tommi Jaakkola.
\newblock Rethinking diffusion models with symmetries through canonicalization
  with applications to molecular graph generation.
\newblock \emph{arXiv preprint arXiv:2602.15022}, 2026.

\bibitem[Morehead and Cheng(2024)]{s42004_024_01233_z}
Alex Morehead and Jianlin Cheng.
\newblock Geometry-complete diffusion for {3D} molecule generation and
  optimization.
\newblock \emph{Communications Chemistry}, 7\penalty0 (1):\penalty0 150, 2024.
\newblock \doi{10.1038/s42004-024-01233-z}.

\bibitem[Vignac et~al.(2023)Vignac, Osman, Toni, and Frossard]{midi}
Clement Vignac, Nagham Osman, Laura Toni, and Pascal Frossard.
\newblock Midi: Mixed graph and 3d denoising diffusion for molecule generation.
\newblock \emph{European Conference on Machine Learning and Principles and
  Practice of Knowledge Discovery in Databases}, 2023.

\bibitem[Song et~al.(2023{\natexlab{a}})Song, Gong, Xu, Cao, Lan, Ermon, Zhou,
  and Ma]{2312_07168v1}
Yuxuan Song, Jingjing Gong, Minkai Xu, Ziyao Cao, Yanyan Lan, Stefano Ermon,
  Hao Zhou, and Wei-Ying Ma.
\newblock Equivariant flow matching with hybrid probability transport for 3d
  molecule generation.
\newblock \emph{Advances in Neural Information Processing Systems},
  36:\penalty0 549--568, 2023{\natexlab{a}}.

\bibitem[Hong et~al.(2024)Hong, Lin, and Tan]{hong2024accelerating}
Haokai Hong, Wanyu Lin, and Kay~Chen Tan.
\newblock Accelerating 3d molecule generation via jointly geometric optimal
  transport.
\newblock \emph{arXiv preprint arXiv:2405.15252}, 2024.

\bibitem[Guan et~al.(2023)Guan, Qian, Peng, Su, Peng, and Ma]{guan20233d}
Jiaqi Guan, Wesley~Wei Qian, Xingang Peng, Yufeng Su, Jian Peng, and Jianzhu
  Ma.
\newblock 3d equivariant diffusion for target-aware molecule generation and
  affinity prediction.
\newblock \emph{arXiv preprint arXiv:2303.03543}, 2023.

\bibitem[Schneuing et~al.(2024)Schneuing, Harris, Du, Didi, Jamasb, Igashov,
  Du, Gomes, Blundell, Lio, et~al.]{schneuing2024structure}
Arne Schneuing, Charles Harris, Yuanqi Du, Kieran Didi, Arian Jamasb, Ilia
  Igashov, Weitao Du, Carla Gomes, Tom~L Blundell, Pietro Lio, et~al.
\newblock Structure-based drug design with equivariant diffusion models.
\newblock \emph{Nature Computational Science}, 4\penalty0 (12):\penalty0
  899--909, 2024.

\bibitem[Rombach et~al.(2022)Rombach, Blattmann, Lorenz, Esser, and
  Ommer]{rombach2022high}
Robin Rombach, Andreas Blattmann, Dominik Lorenz, Patrick Esser, and Bj{\"o}rn
  Ommer.
\newblock High-resolution image synthesis with latent diffusion models.
\newblock In \emph{Proceedings of the IEEE/CVF conference on computer vision
  and pattern recognition}, pages 10684--10695, 2022.

\bibitem[Xu et~al.(2023{\natexlab{b}})Xu, Powers, Dror, Ermon, and
  Leskovec]{2305_01140v1}
Minkai Xu, Alexander Powers, Ron Dror, Stefano Ermon, and Jure Leskovec.
\newblock Geometric latent diffusion models for {3D} molecule generation.
\newblock In \emph{Proceedings of the 40th International Conference on Machine
  Learning}, volume 202 of \emph{Proceedings of Machine Learning Research},
  pages 38592--38610. PMLR, 2023{\natexlab{b}}.

\bibitem[Zhang et~al.(2018)Zhang, Isola, Efros, Shechtman, and
  Wang]{zhang2018unreasonable}
Richard Zhang, Phillip Isola, Alexei~A Efros, Eli Shechtman, and Oliver Wang.
\newblock The unreasonable effectiveness of deep features as a perceptual
  metric.
\newblock In \emph{Proceedings of the IEEE conference on computer vision and
  pattern recognition}, pages 586--595, 2018.

\bibitem[Song et~al.(2023{\natexlab{b}})Song, Dhariwal, Chen, and
  Sutskever]{song2023consistency}
Yang Song, Prafulla Dhariwal, Mark Chen, and Ilya Sutskever.
\newblock Consistency models.
\newblock 2023{\natexlab{b}}.

\bibitem[Mirzaei et~al.(2023)Mirzaei, Aumentado-Armstrong, Derpanis, Kelly,
  Brubaker, Gilitschenski, and Levinshtein]{mirzaei2023spin}
Ashkan Mirzaei, Tristan Aumentado-Armstrong, Konstantinos~G Derpanis, Jonathan
  Kelly, Marcus~A Brubaker, Igor Gilitschenski, and Alex Levinshtein.
\newblock Spin-nerf: Multiview segmentation and perceptual inpainting with
  neural radiance fields.
\newblock In \emph{Proceedings of the IEEE/CVF Conference on Computer Vision
  and Pattern Recognition}, pages 20669--20679, 2023.

\bibitem[Wang et~al.(2025)Wang, Zhou, Gupta, Lin, Jegelka, Bates, and
  Jaakkola]{wang2025learning}
Chenyu Wang, Cai Zhou, Sharut Gupta, Zongyu Lin, Stefanie Jegelka, Stephen
  Bates, and Tommi Jaakkola.
\newblock Learning diffusion models with flexible representation guidance.
\newblock In \emph{ICML 2025 Generative AI and Biology (GenBio) Workshop},
  2025.
\newblock URL \url{https://openreview.net/forum?id=o2W4FTtBVJ}.

\bibitem[Zaidi et~al.(2022)Zaidi, Schaarschmidt, Martens, Kim, Teh,
  Sanchez-Gonzalez, Battaglia, Pascanu, and Godwin]{zaidi2022pre}
Sheheryar Zaidi, Michael Schaarschmidt, James Martens, Hyunjik Kim, Yee~Whye
  Teh, Alvaro Sanchez-Gonzalez, Peter Battaglia, Razvan Pascanu, and Jonathan
  Godwin.
\newblock Pre-training via denoising for molecular property prediction.
\newblock \emph{arXiv preprint arXiv:2206.00133}, 2022.

\bibitem[Perez and Gomez-Bombarelli(2026)]{perez2026self}
Tynan Perez and Rafael Gomez-Bombarelli.
\newblock Self-conditioned denoising for atomistic representation learning.
\newblock \emph{arXiv preprint arXiv:2603.17196}, 2026.

\bibitem[Ramakrishnan et~al.(2014)Ramakrishnan, Dral, Rupp, and von
  Lilienfeld]{ramakrishnan2014quantum}
Raghunathan Ramakrishnan, Pavlo~O Dral, Matthias Rupp, and O~Anatole von
  Lilienfeld.
\newblock Quantum chemistry structures and properties of 134 kilo molecules.
\newblock \emph{Scientific Data}, 1\penalty0 (1):\penalty0 1--7, 2014.

\bibitem[Kingma and Welling(2014)]{kingma2014auto}
Diederik~P Kingma and Max Welling.
\newblock Auto-encoding variational bayes.
\newblock \emph{International Conference on Learning Representations}, 2014.

\bibitem[Bengio et~al.(2013)Bengio, L{\'e}onard, and
  Courville]{bengio2013estimating}
Yoshua Bengio, Nicholas L{\'e}onard, and Aaron Courville.
\newblock Estimating or propagating gradients through stochastic neurons for
  conditional computation.
\newblock \emph{arXiv preprint arXiv:1308.3432}, 2013.

\bibitem[Axelrod and Gomez-Bombarelli(2022)]{axelrod2022geom}
Simon Axelrod and Rafael Gomez-Bombarelli.
\newblock Geom, energy-annotated molecular conformations for property
  prediction and molecular generation.
\newblock \emph{Scientific Data}, 9\penalty0 (1):\penalty0 185, 2022.

\bibitem[Irwin et~al.(2024)Irwin, Tibo, Janet, and Olsson]{irwin2024semlaflow}
Ross Irwin, Alessandro Tibo, Jon~Paul Janet, and Simon Olsson.
\newblock Semlaflow--efficient 3d molecular generation with latent attention
  and equivariant flow matching.
\newblock \emph{arXiv preprint arXiv:2406.07266}, 2024.

\bibitem[Dunn and Koes(2024)]{flowmol}
Ian Dunn and David~Ryan Koes.
\newblock Mixed continuous and categorical flow matching for 3d de novo
  molecule generation.
\newblock \emph{ArXiv}, pages arXiv--2404, 2024.

\bibitem[Huang et~al.(2023)Huang, Sun, Du, and Lv]{jodo}
Han Huang, Leilei Sun, Bowen Du, and Weifeng Lv.
\newblock Learning joint 2d \& 3d diffusion models for complete molecule
  generation.
\newblock \emph{arXiv preprint arXiv:2305.12347}, 2023.

\bibitem[Le et~al.(2023)Le, Cremer, Noe, Clevert, and Sch{\"u}tt]{eqgatdiff}
Tuan Le, Julian Cremer, Frank Noe, Djork-Arn{\'e} Clevert, and Kristof
  Sch{\"u}tt.
\newblock Navigating the design space of equivariant diffusion-based generative
  models for de novo 3d molecule generation.
\newblock \emph{arXiv preprint arXiv:2309.17296}, 2023.

\bibitem[Wu et~al.(2017)Wu, Ramsundar, Feinberg, Gomes, Geniesse, Pappu,
  Leswing, and Pande]{wu2017moleculenet}
Zhenqin Wu, Bharath Ramsundar, Evan~N. Feinberg, Joseph Gomes, Caleb Geniesse,
  Aneesh~S. Pappu, Karl Leswing, and Vijay Pande.
\newblock Moleculenet: A benchmark for molecular machine learning.
\newblock \emph{arXiv preprint arXiv:1703.00564}, 2017.
\newblock URL \url{https://arxiv.org/abs/1703.00564}.

\bibitem[Roy and Vetterli(2007)]{roy2007effective}
Olivier Roy and Martin Vetterli.
\newblock The effective rank: A measure of effective dimensionality.
\newblock In \emph{2007 15th European signal processing conference}, pages
  606--610. IEEE, 2007.

\end{thebibliography}

\newpage

\appendix
\startcontents[appendix]

\section*{Appendix Contents}
\thispagestyle{empty}
\addcontentsline{toc}{section}{Appendix Contents}
\printcontents[appendix]{}{1}{}
\clearpage

\section{Methods and Algorithms}
\label{sec:appendix-methods}

\subsection{Additional Implementation Details}
\label{sec:method-details}

In this section, we provide additional formulation details for \modelname{}, including the representation head, the molecule perceptual loss, the representation projection alignment loss, and the overall training objective.

\paragraph{Molecular Representation and Encoder Outputs.}
A 3D molecule is represented as $\mathcal{M}=(\bm{x},\bm{a},\mathcal{E})$, where $\bm{x}\in\mathbb{R}^{N\times 3}$ denotes atomic coordinates, $\bm{a}$ denotes atom types, and $\mathcal{E}$ denotes chemical bonds. We additionally use an atom mask $\bm{m}\in\{0,1\}^{N}$ to distinguish valid atoms from padding atoms, where $m_i=1$ indicates that atom $i$ is valid.

Given a molecule $\mathcal{M}$, the frozen pretrained encoder extracts layer-wise representations. For the graph-level representation head, we denote the graph-level output from the $l$-th encoder layer as
\begin{equation}
    \bm{H}^{(l)}(\mathcal{M}) \in \mathbb{R}^{d},
    \qquad l=1,\ldots,L,
\end{equation}
where $L$ is the number of encoder layers and $d$ is the graph representation dimension. In practice, these graph-level representations are obtained from atom-level hidden states through masked pooling over valid atoms, followed by normalization.

\paragraph{Representation Head.}
To obtain a more structured and generation-friendly latent space, we introduce a representation head on top of the multi-layer graph representations of the pretrained encoder. Specifically, we aggregate layer-wise representations through learnable layer pooling:
\begin{equation}
    \bm{g}(\mathcal{M})
    =
    \sum_{l=1}^{L}\alpha_l \bm{H}^{(l)}(\mathcal{M}),
    \qquad
    \alpha_l=\textsc{Softmax}(\bm{w})_l,
\end{equation}
where $\bm{g}(\mathcal{M})\in\mathbb{R}^{d}$ is the pooled graph representation, $\bm{w}\in\mathbb{R}^{L}$ denotes the learnable layer-weight logits, and $\alpha_l$ is the normalized importance weight assigned to the $l$-th encoder layer.

The pooled representation is then mapped to the parameters of a diagonal Gaussian posterior:
\begin{equation}
    \boldsymbol{\mu}
    =
    f_\mu(\bm{g}(\mathcal{M})),
    \qquad
    \log\boldsymbol{\sigma}^{2}
    =
    \mathrm{clamp}
    \left(
    f_{\log\sigma^2}(\bm{g}(\mathcal{M})),
    v_{\min},
    v_{\max}
    \right),
\end{equation}
where $f_\mu$ and $f_{\log\sigma^2}$ are learnable projection heads for predicting the mean and log-variance, respectively. The clamp operation constrains the predicted log-variance within $[v_{\min},v_{\max}]$ for numerical stability. The latent representation is sampled via the reparameterization trick:
\begin{equation}
    \bm{z}
    =
    \boldsymbol{\mu}
    +
    \boldsymbol{\sigma}\odot\boldsymbol{\epsilon},
    \qquad
    \boldsymbol{\epsilon}\sim\mathcal{N}(\bm{0},\bm{I}),
\end{equation}
where $\odot$ denotes element-wise multiplication. The sampled latent code $\bm{z}$ is then used as the conditioning representation for molecule generation.

The representation head is trained jointly with the generator. On the one hand, the generation loss encourages the representation head to extract molecule representations that are useful for reconstructing clean molecular structures from noisy states. On the other hand, we regularize the posterior with a KL divergence toward a standard Gaussian prior:
\begin{equation}
    \mathcal{L}_{\textsc{KL}}
    =
    -\frac{1}{2}
    \sum_{j=1}^{d_z}
    \left(
    1+\log\sigma_j^2-\mu_j^2-\sigma_j^2
    \right),
    \label{eq:kl-detail}
\end{equation}
where $d_z$ is the latent dimension. This regularization encourages a compact and smooth latent representation space.

\paragraph{Straight-Through Estimator for Discrete Atom Types.}
The molecule perceptual loss requires feeding the generated molecule $\hat{\mathcal{M}}$ back into the frozen pretrained encoder. Since the encoder takes discrete atom types as input, while the generator predicts continuous atom-type logits, we use a straight-through estimator to construct differentiable atom-type inputs:
\begin{equation}
    \hat{\bm{a}}_{\mathrm{ST}}
    =
    \mathrm{onehot}
    \left(
    \arg\max \hat{\bm{a}}_{\mathrm{logits}}
    \right)
    +
    \hat{\bm{a}}_{\mathrm{logits}}
    -
    \mathrm{sg}
    \left(
    \hat{\bm{a}}_{\mathrm{logits}}
    \right),
\end{equation}
where $\hat{\bm{a}}_{\mathrm{logits}}$ denotes the predicted atom-type logits and $\mathrm{sg}(\cdot)$ denotes the stop-gradient operation. This keeps the forward input discrete while allowing gradients to flow through the continuous logits during backpropagation.

\paragraph{Molecule Perceptual Loss.}
During training, the generator predicts a denoised molecule $\hat{\mathcal{M}}$ from a noisy input. We pass both the clean molecule $\mathcal{M}$ and the denoised molecule $\hat{\mathcal{M}}$ through the frozen pretrained encoder, and compute their pooled graph-level representations using the same learnable layer-pooling module as in the representation head. The molecule perceptual loss is defined as
\begin{equation}
    \mathcal{L}_{\textsc{perc}}
    =
    w(t)
    \left\|
    \mathrm{sg}
    \left(
    \bm{g}(\mathcal{M})
    \right)
    -
    \bm{g}(\hat{\mathcal{M}})
    \right\|_2^2,
    \label{eq:perc-detail}
\end{equation}
where $\bm{g}(\cdot)$ denotes the pooled graph-level representation extracted by the frozen encoder, and $w(t)$ is a time-dependent weighting function. The clean-molecule representation is treated as a fixed semantic target, while gradients are allowed to flow through $\bm{g}(\hat{\mathcal{M}})$ to optimize the generator. Although the encoder parameters are frozen, gradients can still be backpropagated through the encoder computation graph to the generated molecular coordinates and atom-type logits.

We implement $w(t)$ as a cosine schedule that assigns larger weights to later denoising timesteps, where the generated molecule is closer to clean data and the encoder representations provide a more informative semantic alignment signal. Concretely, with normalized diffusion time $t\in[0,1]$, we use
\begin{equation}
    w(t)
    =
    w_{\min}
    +
    (w_{\max}-w_{\min})
    \cdot
    \frac{1+\cos(\pi t)}{2},
    \label{eq:cosine-weight-detail}
\end{equation}
where $w_{\min}$ and $w_{\max}$ control the minimum and maximum alignment weights.

\paragraph{Representation Projection Alignment Loss.}
Beyond graph-level perceptual alignment, we further align the internal node-level features of the generator with the pretrained encoder representations. Let
\begin{equation}
    \bm{H}_{\mathrm{enc}}^{(l_e)}
    \in
    \mathbb{R}^{N\times d_e},
    \qquad
    \bm{H}_{\mathrm{gen}}^{(l_g)}
    \in
    \mathbb{R}^{N\times d_g}
\end{equation}
denote node features from the $l_e$-th encoder layer and the $l_g$-th generator layer, respectively. Here, $d_e$ and $d_g$ are the corresponding feature dimensions. To reconcile dimension and scale differences, we use a learnable projection
\begin{equation}
    \textsc{Proj}:\mathbb{R}^{d_g}\rightarrow\mathbb{R}^{d_e}
\end{equation}
to map generator features into the encoder feature space.

The representation projection alignment loss is defined over the set of valid atoms $\mathcal{A}=\{i:m_i=1\}$:
\begin{equation}
    \mathcal{L}_{\textsc{repa}}
    =
    -
    \frac{1}{|\mathcal{A}|}
    \sum_{i\in\mathcal{A}}
    \left\langle
    \frac{
    \textsc{Proj}
    \left(
    \bm{H}_{\mathrm{gen},i}^{(l_g)}
    \right)
    }{
    \left\|
    \textsc{Proj}
    \left(
    \bm{H}_{\mathrm{gen},i}^{(l_g)}
    \right)
    \right\|_2
    },
    \frac{\mathrm{sg}
    \left(
    \hat{\bm{H}}_{\mathrm{enc},i}^{(l_e)}\right)
    }{
    \left\|
    \mathrm{sg}
    \left(\hat{\bm{H}}_{\mathrm{enc},i}^{(l_e)}\right)
    \right\|_2
    }
    \right\rangle,
    \label{eq:repa-detail}
\end{equation}
where $\langle\cdot,\cdot\rangle$ denotes the inner product. We use cosine similarity rather than MSE to reduce the influence of feature-scale discrepancies across the encoder and generator representation spaces~\citep{yu2024repa}.

\paragraph{Total Training Objective.}
Combining the generation objective with the proposed representation-level regularization terms gives the full Phase-I training objective:
\begin{equation}
    \mathcal{L}_{\mathrm{total}}
    =
    \mathcal{L}_{\mathrm{gen}}
    +
    \lambda_{\textsc{KL}}\mathcal{L}_{\textsc{KL}}
    +
    \lambda_{\textsc{perc}}\mathcal{L}_{\textsc{perc}}
    +
    \lambda_{\textsc{repa}}\mathcal{L}_{\textsc{repa}},
    \label{eq:total-detail}
\end{equation}
where $\mathcal{L}_{\mathrm{gen}}$ is the base molecule generation loss determined by the chosen generative backbone, and $\lambda_{\textsc{KL}}$, $\lambda_{\textsc{perc}}$, and $\lambda_{\textsc{repa}}$ are loss-weight hyperparameters.

\paragraph{Inference.}
At inference time, generation follows the two-stage GeoRCG pipeline~\citep{li2024geometric}. First, the representation diffusion model samples a latent representation $\hat{\boldsymbol{\mu}}\sim p_\phi$, optionally conditioned on atom count, molecular properties, or other constraints. Second, the molecule generator produces a 3D molecular structure conditioned on the sampled representation:
\begin{equation}
    \hat{\mathcal{M}}
    \sim
    p_\theta
    \left(
    \cdot
    \mid
    \hat{\boldsymbol{\mu}}
    \right).
\end{equation}
This decoupled design allows latent representation generation and 3D molecular structure generation to be modeled separately.
\subsection{Training Algorithm}
\label{sec:training-algorithm}

Algorithm~\ref{alg:training} summarizes the Phase-I training procedure of \modelname{} for the representation-conditioned molecular generator.

\begin{algorithm}[h]
\caption{\modelname{} Phase-I Training}
\label{alg:training}
\begin{algorithmic}[1]
\REQUIRE Frozen pretrained encoder $f_{\mathrm{enc}}$, generator $p_\theta$, representation head $q_\psi$, training set $\mathcal{D}$
\FOR{each training iteration}
    \STATE Sample a clean molecule $\mathcal{M}=(\bm{x},\bm{a},\mathcal{E})\sim\mathcal{D}$
    \STATE Extract multi-layer graph representations $\{\bm{H}^{(l)}(\mathcal{M})\}_{l=1}^{L}$ using $f_{\mathrm{enc}}$
    \STATE Compute pooled representation $\bm{g}(\mathcal{M})=\sum_{l=1}^{L}\alpha_l\bm{H}^{(l)}(\mathcal{M})$
    \STATE Predict posterior parameters $(\boldsymbol{\mu},\log\boldsymbol{\sigma}^{2}) \leftarrow q_\psi(\bm{g}(\mathcal{M}))$
    \STATE Sample latent code $\bm{z}=\boldsymbol{\mu}+\boldsymbol{\sigma}\odot\boldsymbol{\epsilon}$, where $\boldsymbol{\epsilon}\sim\mathcal{N}(\bm{0},\bm{I})$
    \STATE Generate a noisy molecular state and compute the backbone generation loss $\mathcal{L}_{\mathrm{gen}}$ conditioned on $\bm{z}$
    \STATE Obtain the denoised molecule prediction $\hat{\mathcal{M}}$ from the generator
    \STATE Feed $\hat{\mathcal{M}}$ into $f_{\mathrm{enc}}$ using the straight-through estimator for predicted atom types
    \STATE Compute the molecule perceptual loss $\mathcal{L}_{\textsc{perc}}$ between $\bm{g}(\hat{\mathcal{M}})$ and $\mathrm{sg}(\bm{g}(\mathcal{M}))$
    \STATE Extract encoder node features $\bm{H}_{\mathrm{enc}}^{(l_e)}$ and generator node features $\bm{H}_{\mathrm{gen}}^{(l_g)}$
    \STATE Map $\bm{H}_{\mathrm{gen}}^{(l_g)}$ into encoder feature space with $\textsc{Proj}$ 
    \STATE Compute the representation projection alignment loss $\mathcal{L}_{\textsc{repa}}$
    \STATE Compute $\mathcal{L}_{\textsc{KL}}$ from $(\boldsymbol{\mu},\boldsymbol{\sigma})$
    \STATE Compute $\mathcal{L}_{\mathrm{total}}$ according to Eq.~\ref{eq:total-detail}
    \STATE Update $\theta, \psi$ and $\textsc{Proj}$ by gradient descent on $\mathcal{L}_{\mathrm{total}}$, keeping $f_{\mathrm{enc}}$ frozen
\ENDFOR
\end{algorithmic}
\end{algorithm}

\subsection{Auxiliary Metric Algorithms}
\label{sec:appendix-metric-algorithms}

This section provides algorithmic details for the auxiliary metrics used in the representation quality analysis.

\subsubsection{Effective Rank}
\label{sec:effective-rank}

The {effective rank}~\citep{roy2007effective} measures how many
dimensions are effectively used by a representation matrix
$\mathbf{Z} \in \mathbb{R}^{N \times D}$, where $N$ is the number of samples
and $D$ is the representation dimensionality. We compute it from the
normalized eigenvalue spectrum of the representation covariance matrix. In
practice, this is equivalently obtained from the squared singular values of the
centered representation matrix.

\bigskip
\begin{algorithm}[H]
\caption{Effective Rank}
\label{alg:eff-rank}
\begin{algorithmic}[1]
\REQUIRE Representation matrix $\mathbf{Z} \in \mathbb{R}^{N \times D}$
\ENSURE Effective rank $\mathrm{er} \in \mathbb{R}$
\STATE Center the representation matrix:
$
\bar{\mathbf{Z}} \gets \mathbf{Z} - \mathbf{1}\mathbf{1}^{\top}\mathbf{Z}/N
$
\STATE Compute singular values $\mathbf{s} = (\sigma_1, \ldots, \sigma_r)$ of $\bar{\mathbf{Z}}$ via SVD, where $r=\min(N,D)$
\STATE Normalize the squared singular-value spectrum:
$
p_i \gets {\sigma_i^2}/{\sum_{j=1}^{r}\sigma_j^2}
$
\STATE Filter out near-zero values: keep $p_i > \epsilon$ with $\epsilon = 10^{-10}$
\STATE Compute Shannon entropy:
$
H(\mathbf{p}) = -\sum_i p_i \log p_i$
\RETURN $\exp(H(\mathbf{p}))$
\end{algorithmic}
\end{algorithm}

\subsubsection{Empirical Lipschitz Constant}
\label{alg:empirical-lipschitz}

The \textbf{empirical Lipschitz constant} measures how much the representation
output changes in response to small input perturbations.  For a representation
mapping $f: \mathbb{R}^{N \times 3} \to \mathbb{R}^D$ (molecule coordinates
$\to$ latent vector) we estimate:

\begin{equation}
    L_{\text{emp}} = \max_i \frac{\| f(\mathbf{X}_i + \boldsymbol{\epsilon}_i)
      - f(\mathbf{X}_i) \|_2}{\| \boldsymbol{\epsilon}_i \|_2}
\end{equation}

where $\boldsymbol{\epsilon}_i$ are small random coordinate perturbations
applied to each molecule $i$.

\bigskip
\begin{algorithm}[H]
\caption{Empirical Lipschitz Constant}
\label{alg:emp-lipschitz}
\begin{algorithmic}[1]
\REQUIRE Encoder $E$, molecules $\mathbf{C} \in \mathbb{R}^{N \times 3}$, radius $\epsilon$, num samples $S$
\ENSURE Max and mean Lipschitz constants $(L_{\max}, L_{\text{mean}})$
\STATE $L_{\max} \gets 0$, $L_{\text{sum}} \gets 0$
\FOR{$i = 1$ \TO $S$}
  \STATE $\mathbf{c}_i \gets \mathbf{C}[i]$
  \STATE Draw $\boldsymbol{\delta} \sim \mathcal{N}(\mathbf{0}, \mathbf{I})$, normalize to radius $\epsilon$
  \STATE $\mathbf{c}_i' \gets \mathbf{c}_i + \boldsymbol{\delta}$
  \STATE Compute representations: $\mathbf{z}_0 \gets E(\mathbf{c}_i)$, $\mathbf{z}_{\text{pert}} \gets E(\mathbf{c}_i')$
  \STATE $d_z \gets \| \mathbf{z}_{\text{pert}} - \mathbf{z}_0 \|_2 / \epsilon$
  \STATE $L_{\max} \gets \max(L_{\max}, d_z)$
  \STATE $L_{\text{sum}} \gets L_{\text{sum}} + d_z$
\ENDFOR
\STATE $L_{\text{mean}} \gets L_{\text{sum}} / S$
\RETURN $(L_{\max}, L_{\text{mean}})$
\end{algorithmic}
\end{algorithm}

\textbf{Note:} We use $\epsilon = 10^{-3}$ and $S = 100$ perturbations.
A \emph{smaller} empirical Lipschitz constant indicates a \emph{smoother}
representation mapping.

\subsubsection{Pairwise Distance Distribution}
\label{sec:pairwise-distance}

To characterize the geometric structure of the representation space, we
compute all pairwise Euclidean distances between samples:

\bigskip
\begin{algorithm}[H]
\caption{Pairwise Distance Statistics}
\label{alg:pair-dist}
\begin{algorithmic}[1]
\REQUIRE Representations $\mathbf{Z} \in \mathbb{R}^{N \times D}$
\ENSURE Statistics $(\mu_D, \sigma_D, \text{median}_D, p_{10}, p_{90})$
\STATE Compute all pairwise distances: $D_{ij} = \| \mathbf{z}_i - \mathbf{z}_j \|_2 \quad \forall i < j$
\STATE $N_{\text{pair}} \gets N(N-1)/2$
\STATE $\mu_D \gets \frac{1}{N_{\text{pair}}} \sum_k D_k$
\STATE $\sigma_D \gets \sqrt{\frac{1}{N_{\text{pair}}} \sum_k (D_k - \mu_D)^2}$
\STATE $\text{median}_D \gets \text{median}(D_k)$
\STATE $p_{10} \gets \text{percentile}(D, 10)$
\STATE $p_{90} \gets \text{percentile}(D, 90)$
\RETURN $(\mu_D, \sigma_D, \text{median}_D, p_{10}, p_{90})$
\end{algorithmic}
\end{algorithm}

\bigskip\noindent
\textbf{Interpretation:}
A \emph{low} mean and std indicates the representation is compressed into a small region of space; a \emph{larger} mean and std indicates a well-spread representation space. Values for both representation spaces are reported in \Cref{tab:appendix-summary-m}.

\subsubsection{k-Nearest Neighbor Smoothness}
\label{sec:knn-smoothness}

Local smoothness measures how close each sample's neighbors are in the
representation space:

\bigskip
\begin{algorithm}[H]
\caption{k-NN Average Distance (Local Smoothness)}
\label{alg:knn-smooth}
\begin{algorithmic}[1]
\REQUIRE Representations $\mathbf{Z} \in \mathbb{R}^{N \times D}$, neighbor count $k$
\ENSURE Local smoothness score
\STATE Fit $k$-NN model on $\mathbf{Z}$ with Euclidean metric
\FOR{$i = 1$ \TO $N$}
  \STATE Find $k$-nearest neighbors (excluding self): $N_k(i) = \{j_1, \ldots, j_k\}$
  \STATE $d_i \gets \frac{1}{k} \sum_{j \in N_k(i)} \| \mathbf{z}_i - \mathbf{z}_j \|_2$
\ENDFOR
\STATE $\text{smoothness} \gets \frac{1}{N} \sum_i d_i$
\RETURN $\text{smoothness}$
\end{algorithmic}
\end{algorithm}

\textbf{Note:} We exclude the sample itself from its own neighborhood
(use neighbors $k+1$, then discard self).

\section{Experimental Results}
\label{sec:appendix-results}

\subsection{Layer Probe Analysis: Complete Results}
\label{sec:layer-probe-appendix}

This section provides the complete linear probe results for all 10 functional
groups across all 9 encoder layers, supplementing the summary presented in
\Cref{fig:layer-analysis}a.

\subsubsection{Experimental Setup}
\label{sec:probe-setup}

We follow the methodology described in \autoref{sec:motivation}.
The Frad encoder~\citep{frad} (Equivariant Transformer, 9 layers, 256-dimensional
representations) is applied to 30,000 QM9 molecules in a frozen state.
For each layer $l \in \{0, 1, \dots, 8\}$, a logistic regression probe is
trained on 80\% of the data and evaluated on the held-out 20\% (stratified split).
We report balanced accuracy to handle class imbalance. See \Cref{tab:layer-probe-full} for full results.

\begin{table}[h]
  \centering
  \caption{\textbf{Complete linear probe results: all 10 functional groups $\times$ all 9 layers.}
  Top-1 balanced accuracy ($\%$) on QM9 validation set. The L0 advantage (L0 minus L8) is shown with 95\% confidence intervals from the normal approximation with Bonferroni correction over the 10 comparisons. All differences are significant at $p<0.001$ except fluoro, where both layers are at ceiling.}
  \label{tab:layer-probe-full}
  \resizebox{\textwidth}{!}{
  \begin{tabular}{lcccccccccc}
    \toprule
    \textbf{Layer} & \textbf{Hydroxyl} & \textbf{Amino} & \textbf{Carbonyl} & \textbf{Carboxyl} & \textbf{Cyano} & \textbf{Ether} & \textbf{Fluoro} & \textbf{Alkene} & \textbf{Alkyne} & \textbf{Aromatic} \\
    \midrule
    L0   & 99.8 & 97.4 & 99.4 & 98.2 & 99.98 & 99.9 & 100.0 & 96.4 & 99.99 & 98.9 \\
    L1          & 99.5 & 96.4 & 99.3 & 98.0 & 99.97 & 99.8 & 100.0 & 96.1 & 99.99 & 98.6 \\
    L2          & 99.1 & 95.4 & 99.2 & 97.8 & 99.96 & 99.8 & 100.0 & 95.8 & 99.98 & 98.5 \\
    L3          & 98.5 & 94.1 & 99.1 & 97.5 & 99.95 & 99.7 & 100.0 & 95.5 & 99.98 & 98.3 \\
    L4          & 97.4 & 92.3 & 98.8 & 97.1 & 99.94 & 99.6 & 100.0 & 94.8 & 99.97 & 98.1 \\
    L5          & 96.3 & 90.3 & 98.4 & 96.8 & 99.93 & 99.2 & 100.0 & 94.2 & 99.96 & 97.8 \\
    L6          & 95.1 & 87.6 & 97.8 & 96.5 & 99.92 & 98.5 & 100.0 & 93.7 & 99.95 & 97.6 \\
    L7          & 93.9 & 85.3 & 97.3 & 96.4 & 99.91 & 97.2 & 100.0 & 93.2 & 99.94 & 97.6 \\
    L8    & 92.6 & 82.9 & 96.1 & 96.4 & 99.62 & 85.8 & 100.0 & 92.5 & 99.31 & 97.7 \\
    \midrule
    \textbf{L0 $-$ L8} & $+7.2_{\pm0.3}$ & $+14.5_{\pm0.5}$ & $+3.3_{\pm0.2}$ & $+1.8_{\pm0.2}$ & $+0.4_{\pm0.0}$ & $+14.1_{\pm0.3}$ & $0.0$ & $+3.9_{\pm0.3}$ & $+0.7_{\pm0.1}$ & $+1.2_{\pm0.2}$ \\
    \midrule
    \textbf{$n_{\text{pos}}$} & 10,016 & 9,332 & 8,886 & 142 & 3,278 & 19,088 & 354 & 2,926 & 3,747 & 9,563 \\
    \bottomrule
  \end{tabular}}
\end{table}

\subsubsection{PCA Variance Decomposition per Layer}
\label{sec:probe-pca-detail}

\autoref{tab:pca-variance} supplements the analysis in
\autoref{sec:motivation} by reporting the per-component PCA explained variance
ratio for each layer.

Early layers concentrate variance in a low-dimensional subspace: the top five principal components of L0 explain $83.3\%$ of the total variance, against $45.0\%$ at L8. Effective rank rises nearly monotonically from $3.1$ at L0 to $6.2$ at L8. Deeper layers thus encode molecular information across a higher-dimensional subspace, whereas early layers keep functional-group information in a concentrated and linearly separable one, which is what the probes in \Cref{tab:layer-probe-full} measure.

\begin{table}[h]
  \centering
  \caption{\textbf{PCA explained variance ratio per layer.}
  Each entry shows the fraction of total variance captured by the top-$k$
  principal components ($k=1,3,5$). Variance is concentrated in early layers and becomes progressively more distributed in later layers. Effective rank is $\exp(H(\mathbf{p}))$ as defined in \Cref{sec:effective-rank}.}
  \label{tab:pca-variance}
  \begin{tabular}{lcccccc}
    \toprule
    \textbf{Layer} & \textbf{PC1} & \textbf{PC2} & \textbf{PC3} & \textbf{Top-3} & \textbf{Top-5} & \textbf{Eff.\ Rank} \\
    \midrule
    L0 & 31.0\% & 23.5\% & 14.3\% & 68.8\% & 83.3\% & 3.1 \\
    L1 & 23.2\% & 19.4\% & 14.1\% & 56.7\% & 71.4\% & 3.8 \\
    L2 & 35.0\% & 17.8\% & 10.2\% & 63.0\% & 71.5\% & 3.4 \\
    L3 & 28.5\% & 15.9\% &  9.6\% & 54.0\% & 62.0\% & 4.1 \\
    L4 & 23.2\% & 15.6\% &  9.4\% & 48.2\% & 56.7\% & 4.6 \\
    L5 & 23.1\% & 15.3\% &  8.8\% & 47.2\% & 54.6\% & 4.8 \\
    L6 & 25.7\% & 15.1\% &  9.7\% & 50.5\% & 61.2\% & 4.4 \\
    L7 & 16.3\% & 14.8\% &  9.3\% & 40.4\% & 49.9\% & 5.5 \\
    L8 & 13.6\% & 13.2\% &  9.1\% & 35.9\% & 45.0\% & 6.2 \\
    \bottomrule
  \end{tabular}
\end{table}
\subsubsection{Cross-Layer Correlation Matrix}
\label{sec:probe-correlation-detail}

\autoref{tab:cross-layer-corr} reports the full pairwise cosine similarity
matrix between layer-level mean representations, confirming the two-block structure
identified in \autoref{sec:motivation}.

\begin{table}[h]
  \centering
  \caption{\textbf{Cross-layer cosine similarity matrix.}
  Each entry $(i,j)$ shows the cosine similarity between the mean graph
  representation of layer $i$ and layer $j$. Two blocks are clearly visible:
  L0--L5 (feature extraction, $r=0.66$--$0.95$) and L7--L8 (semantic aggregation,
  $r=0.96$). The sharpest transition occurs at L5$\rightarrow$L6.}
  \label{tab:cross-layer-corr}
  \begin{tabular}{l|ccccccccc}
    \toprule
              & \textbf{L0} & \textbf{L1} & \textbf{L2} & \textbf{L3} & \textbf{L4} & \textbf{L5} & \textbf{L6} & \textbf{L7} & \textbf{L8} \\
    \midrule
    L0 & 1.00 & 0.95 & 0.89 & 0.83 & 0.69 & 0.66 & 0.51 & 0.54 & 0.52 \\
    L1 & 0.95 & 1.00 & 0.95 & 0.89 & 0.77 & 0.72 & 0.54 & 0.54 & 0.52 \\
    L2 & 0.89 & 0.95 & 1.00 & 0.93 & 0.83 & 0.80 & 0.60 & 0.53 & 0.52 \\
    L3 & 0.83 & 0.89 & 0.93 & 1.00 & 0.90 & 0.85 & 0.64 & 0.54 & 0.51 \\
    L4 & 0.69 & 0.77 & 0.83 & 0.90 & 1.00 & 0.94 & 0.72 & 0.52 & 0.48 \\
    L5 & 0.66 & 0.72 & 0.80 & 0.85 & 0.94 & 1.00 & 0.77 & 0.52 & 0.50 \\
    L6 & 0.51 & 0.54 & 0.60 & 0.64 & 0.72 & 0.77 & 1.00 & 0.61 & 0.61 \\
    L7 & 0.54 & 0.54 & 0.53 & 0.54 & 0.52 & 0.52 & 0.61 & 1.00 & 0.96 \\
    L8 & 0.52 & 0.52 & 0.52 & 0.51 & 0.48 & 0.50 & 0.61 & 0.96 & 1.00 \\
    \bottomrule
  \end{tabular}
\end{table}

\begin{figure}[h]
  \centering
  \includegraphics[width=0.75\linewidth]{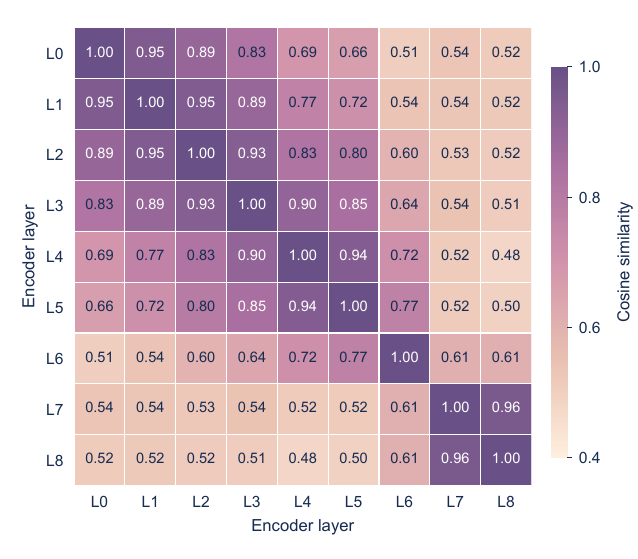}
  \caption{\textbf{Cross-layer cosine similarity heatmap.}
    Pairwise cosine similarity between mean graph-level representations
    from nine Frad encoder layers on QM9.
    }
  \label{fig:layer-probe-full}
\end{figure}

\subsection{Representation Smoothness Analysis: Supplementary Figures}
\label{sec:smoothness-appendix}

This section supplements the representation quality analysis in
\autoref{sec:exp_rep_quality} with additional visualizations of the representation head's
latent space on GEOM-DRUG~\citep{axelrod2022geom}.

\subsubsection{Pairwise Distance Summary}
\label{sec:appendix-pairwise}

\begin{figure}[h]
  \centering
  \includegraphics[width=0.85\linewidth]{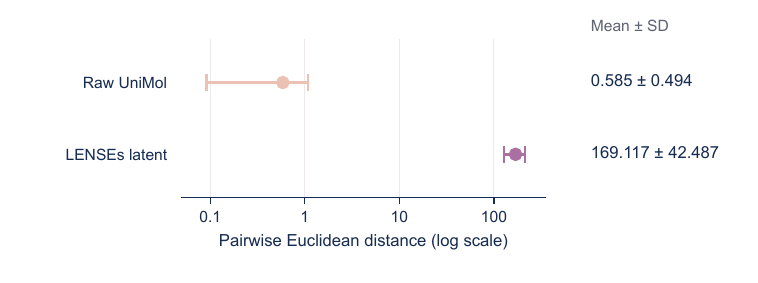}
  \caption{\textbf{Pairwise distance summary: raw encoder vs.\ representation latent space (GEOM-DRUG).}
    Points show mean pairwise Euclidean distances; error bars indicate one standard
    deviation across pairwise distances. The horizontal axis is logarithmic.}
  \label{fig:appendix-pairwise}
\end{figure}

\autoref{fig:appendix-pairwise} summarizes the pairwise Euclidean
distances between molecular representations in the raw UniMol encoder space
versus the representation-head's latent space. The raw encoder representations cluster into
an extremely tight region (mean distance: $0.59$, std: $0.49$). The representation head expands this space by $287\times$
(mean: $169.12$, std: $42.49$), producing a well-structured latent manifold
with meaningful geometric separation between molecules.

\subsection{Representation Quality Metrics}
\label{sec:appendix-rep-quality}

\begin{table}[h]
    \centering
    \caption{\textbf{Representation quality metrics on GEOM-DRUG.}
      }
      \label{tab:appendix-summary-m}
    \resizebox{0.8\textwidth}{!}{
    \begin{tabular}{lccc}
      \toprule
      \textbf{Metric} & \textbf{Raw Encoder} & \textbf{Representation Head's Latent} \\
      \midrule
      Effective Rank                  & 65.32        & \textbf{88.42}     \\
      Lipschitz (mean)      & 2659.81     & \textbf{570.49}      \\
      Lipschitz (max)       & 3545.67     & \textbf{1781.41}     \\
      Pairwise Distance (mean)        & 0.59        & 169.12     \\
      Pairwise Distance (std)         & 0.49        & {42.49}      \\
      k-NN Avg Distance               & 0.19        & {102.94}     \\
      \bottomrule
    \end{tabular}}
\end{table}

\subsection{Subset Ablations}
\label{app:subset_ablations}

To examine the robustness of \modelname{} to its loss weights over a broad hyperparameter range, we conduct a grid ablation study on a randomly sampled $10\%$ subset of the GEOM-DRUG dataset. Each configuration is trained for 300 epochs under the same training protocol. We vary the KL regularization weight and the perceptual weight, and compare the resulting models against the baseline generator trained with the same subset setting. In the tables, cells highlighted in green indicate configurations that outperform the baseline on the corresponding metric.

Table~\ref{tab:subset_baseline_compare} first compares the baseline with two representative variants of \modelname{}. Both variants improve atom stability, molecule stability, validity, energy, and strain. The perceptual-loss variant raises molecule stability from $92.82$ to $97.20$ and validity from $82.86$ to $93.50$, and lowers strain from $111.46$ to $84.19$. The latent refinement therefore improves both chemical validity and stability and the physical quality of generated conformations.

Table~\ref{tab:subset-ablations} reports the full grid over KL and perceptual weights. Of the 180 metric cells in the grid, 135 are better than the baseline. For $\lambda_{\mathrm{KL}}\in[2.5\times10^{-8},5\times10^{-6}]$, molecule stability exceeds the baseline in 24 of 25 configurations and validity in 21 of 25. Within this range the gains therefore do not depend on a single tuned setting. Larger KL weights are examined on the full dataset in \Cref{app:fullset_ablations}.

\begin{table}[htbp]
    \centering
    \caption{
    Representative subset ablations on a randomly sampled $10\%$ subset of GEOM-DRUG.
    Cells highlighted in green indicate results better than the baseline.
    }
    \label{tab:subset_baseline_compare}
    \begin{tabular}{lccc}
    \toprule
    Metric & Baseline & Rep. head KL weight $5\times 10^{-7}$ & Perc. weight 0.001 \\
    \midrule
    atom-stability & 99.76 & \cellcolor{green!20}99.76 & \cellcolor{green!20}99.84 \\
    molecule-stability & 92.82 & \cellcolor{green!20}94.70 & \cellcolor{green!20}97.20 \\
    validity & 82.86 & \cellcolor{green!20}91.00 & \cellcolor{green!20}93.50 \\
    energy & 149.67 & \cellcolor{green!20}105.01 & \cellcolor{green!20}112.01 \\
    strain & 111.46 & \cellcolor{green!20}99.48 & \cellcolor{green!20}84.19 \\
    opt-rmsd & 1.07 & \cellcolor{green!20}0.99 & 1.11 \\
    \bottomrule
    \end{tabular}
\end{table}

\begin{table}[htbp]
    \centering
    \small
    \setlength{\tabcolsep}{4pt}
    \caption{
    Grid ablation of KL regularization weight and perceptual loss weight on a randomly sampled $10\%$ subset of GEOM-DRUG.
    Each configuration is trained for 300 epochs.
    Cells highlighted in green indicate results better than the baseline. REPA weight is fixed to $1\times 10^{-3}$.
    }
    \label{tab:subset-ablations}
    \begin{tabular}{l|lccccc}
    \toprule
    KL weight & Metric & \multicolumn{5}{c}{Perc. Loss weight} \\
    \cmidrule(lr){3-7}
     &  & $5{\times}10^{-4}$ & $7.5{\times}10^{-4}$ & $1{\times}10^{-3}$ & $2.5{\times}10^{-3}$ & $5{\times}10^{-3}$ \\
    \midrule
    \texttt{1e-08} & Energy & \cellcolor{green!20}119.6 & \cellcolor{green!20}136.9 & \cellcolor{green!20}131.3 & 184.0 & 154.4 \\
     & Strain & \cellcolor{green!20}90.77 & \cellcolor{green!20}90.03 & \cellcolor{green!20}91.89 & 135.53 & \cellcolor{green!20}91.67 \\
     & Opt-RMSD & \cellcolor{green!20}0.874 & \cellcolor{green!20}0.990 & 1.171 & 1.285 & 1.191 \\
     & Validity & 74.5 & 37.3 & 73.0 & 71.9 & 79.3 \\
     & Atom Stability & \cellcolor{green!20}99.79 & 98.99 & 99.74 & \cellcolor{green!20}99.78 & \cellcolor{green!20}99.82 \\
     & Mol Stability & \cellcolor{green!20}93.50 & 67.60 & 91.60 & 92.60 & \cellcolor{green!20}94.70 \\
    \midrule
    \texttt{2.5e-08} & Energy & \cellcolor{green!20}107.6 & \cellcolor{green!20}104.3 & \cellcolor{green!20}104.6 & 157.2 & \cellcolor{green!20}143.2 \\
     & Strain & \cellcolor{green!20}80.44 & \cellcolor{green!20}89.82 & \cellcolor{green!20}81.42 & 125.60 & \cellcolor{green!20}92.96 \\
     & Opt-RMSD & 1.105 & 1.144 & 1.085 & 1.164 & 1.143 \\
     & Validity & \cellcolor{green!20}88.1 & \cellcolor{green!20}87.9 & \cellcolor{green!20}85.8 & \cellcolor{green!20}86.3 & 82.3 \\
     & Atom Stability & \cellcolor{green!20}99.83 & \cellcolor{green!20}99.78 & 99.74 & \cellcolor{green!20}99.79 & \cellcolor{green!20}99.82 \\
     & Mol Stability & \cellcolor{green!20}94.80 & \cellcolor{green!20}93.70 & 90.80 & \cellcolor{green!20}93.82 & \cellcolor{green!20}93.62 \\
    \midrule
    \texttt{2.5e-07} & Energy & \cellcolor{green!20}135.9 & \cellcolor{green!20}140.8 & \cellcolor{green!20}122.8 & \cellcolor{green!20}123.8 & \cellcolor{green!20}131.2 \\
     & Strain & \cellcolor{green!20}85.80 & \cellcolor{green!20}83.69 & \cellcolor{green!20}88.98 & \cellcolor{green!20}86.15 & \cellcolor{green!20}94.50 \\
     & Opt-RMSD & \cellcolor{green!20}1.033 & 1.171 & \cellcolor{green!20}0.921 & \cellcolor{green!20}1.002 & \cellcolor{green!20}1.050 \\
     & Validity & \cellcolor{green!20}83.8 & \cellcolor{green!20}89.9 & 80.2 & \cellcolor{green!20}90.5 & \cellcolor{green!20}87.2 \\
     & Atom Stability & \cellcolor{green!20}99.89 & \cellcolor{green!20}99.88 & \cellcolor{green!20}99.80 & \cellcolor{green!20}99.81 & \cellcolor{green!20}99.78 \\
     & Mol Stability & \cellcolor{green!20}97.50 & \cellcolor{green!20}96.40 & \cellcolor{green!20}93.50 & \cellcolor{green!20}94.90 & \cellcolor{green!20}93.20 \\
    \midrule
    \texttt{5e-07} & Energy & \cellcolor{green!20}123.5 & 164.7 & 183.8 & 171.8 & \cellcolor{green!20}114.0 \\
     & Strain & \cellcolor{green!20}105.66 & \cellcolor{green!20}107.00 & 123.39 & \cellcolor{green!20}116.22 & \cellcolor{green!20}83.06 \\
     & Opt-RMSD & 1.141 & \cellcolor{green!20}1.030 & \cellcolor{green!20}0.995 & \cellcolor{green!20}1.041 & 1.093 \\
     & Validity & \cellcolor{green!20}82.9 & \cellcolor{green!20}85.1 & \cellcolor{green!20}87.6 & 79.7 & \cellcolor{green!20}90.9 \\
     & Atom Stability & \cellcolor{green!20}99.80 & \cellcolor{green!20}99.82 & \cellcolor{green!20}99.82 & \cellcolor{green!20}99.80 & \cellcolor{green!20}99.83 \\
     & Mol Stability & \cellcolor{green!20}95.50 & \cellcolor{green!20}95.50 & \cellcolor{green!20}94.20 & \cellcolor{green!20}93.80 & \cellcolor{green!20}96.90 \\
    \midrule
    \texttt{1e-06} & Energy & \cellcolor{green!20}119.9 & \cellcolor{green!20}131.2 & 159.4 & \cellcolor{green!20}124.2 & \cellcolor{green!20}127.3 \\
     & Strain & \cellcolor{green!20}87.24 & \cellcolor{green!20}86.33 & 126.01 & \cellcolor{green!20}89.28 & \cellcolor{green!20}88.72 \\
     & Opt-RMSD & 1.113 & 1.159 & \cellcolor{green!20}1.068 & 1.087 & \cellcolor{green!20}1.062 \\
     & Validity & \cellcolor{green!20}94.0 & \cellcolor{green!20}87.8 & \cellcolor{green!20}83.2 & \cellcolor{green!20}92.7 & \cellcolor{green!20}91.0 \\
     & Atom Stability & \cellcolor{green!20}99.84 & \cellcolor{green!20}99.88 & 99.73 & \cellcolor{green!20}99.89 & \cellcolor{green!20}99.84 \\
     & Mol Stability & \cellcolor{green!20}96.80 & \cellcolor{green!20}97.20 & \cellcolor{green!20}93.00 & \cellcolor{green!20}97.50 & \cellcolor{green!20}96.60 \\
    \midrule
    \texttt{5e-06} & Energy & \cellcolor{green!20}147.3 & 150.4 & \cellcolor{green!20}139.4 & \cellcolor{green!20}142.8 & \cellcolor{green!20}140.8 \\
     & Strain & \cellcolor{green!20}101.82 & \cellcolor{green!20}91.43 & \cellcolor{green!20}105.54 & \cellcolor{green!20}98.00 & \cellcolor{green!20}90.29 \\
     & Opt-RMSD & \cellcolor{green!20}1.036 & 1.071 & \cellcolor{green!20}1.070 & 1.116 & \cellcolor{green!20}1.010 \\
     & Validity & \cellcolor{green!20}88.0 & \cellcolor{green!20}84.5 & \cellcolor{green!20}89.4 & 80.8 & \cellcolor{green!20}83.2 \\
     & Atom Stability & \cellcolor{green!20}99.85 & \cellcolor{green!20}99.86 & \cellcolor{green!20}99.85 & \cellcolor{green!20}99.91 & \cellcolor{green!20}99.83 \\
     & Mol Stability & \cellcolor{green!20}96.40 & \cellcolor{green!20}96.60 & \cellcolor{green!20}96.70 & \cellcolor{green!20}96.70 & \cellcolor{green!20}95.80 \\
    \bottomrule
    \end{tabular}
\end{table}

\subsection{Full Metric of Fullset Ablations}
\label{app:fullset_ablations}

To provide a more complete view of the effect of each auxiliary objective, we report the full ablation results over the loss weights in \Cref{tab:full-ablation-grid}. The REPA rows cover $10^{-6}$ to $10^{-2}$; a finer sweep around $10^{-3}$ is given in the main-text \Cref{tab:main-ablation}. With $\lambda_{\mathrm{KL}}=5\times10^{-7}$ fixed, every REPA weight across four orders of magnitude ($10^{-6}$--$10^{-2}$) and every perceptual weight across a $20\times$ range ($5\times10^{-4}$--$10^{-2}$) lowers both energy and strain relative to the baseline. The KL weight is stable across nearly two orders of magnitude ($10^{-8}$--$5\times10^{-7}$); larger weights pull the posterior toward the prior, as is typical for KL regularization, and the representation collapses at $10^{-4}$.

\begin{sidewaystable}[t]
    \centering
    \caption{
    Full loss-weight ablations on the full GEOM-DRUG set, covering the KL, perceptual, and REPA losses. For the perceptual and REPA ablations we fix $\lambda_{\mathrm{KL}}=5\times10^{-7}$. Percentage-based metrics are reported in percent. All models in this table are trained under exactly the same protocol, except for the ablated hyperparameters.}
    \vspace{1em}
    \label{tab:full-ablation-grid}
    \resizebox{\textheight}{!}{
    \begin{tabular}{lccccccccccccccc}
        \toprule
        Setting
        & $\lambda_{\mathrm{KL}}$
        & $\lambda_{\mathrm{perc}}$
        & $\lambda_{\mathrm{REPA}}$
        & Atom Stable ($\%$) $\uparrow$
        & Mol. Stable ($\%$) $\uparrow$
        & Validity ($\%$) $\uparrow$
        & Unique ($\%$) $\uparrow$
        & Novel ($\%$) $\uparrow$
        & Energy $\downarrow$
        & Energy/Atom $\downarrow$
        & Energy Valid ($\%$) $\uparrow$
        & Opt. Energy Valid ($\%$) $\uparrow$
        & Opt. RMSD $\downarrow$
        & Strain $\downarrow$
        & Strain/Atom $\downarrow$ \\
        \midrule

        Baseline
        & -- & -- & --
        & 99.85 & 97.20 & 94.20 & 100.00 & 99.58
        & 103.33 & 2.2978 & 94.10 & 94.10
        & 0.8730 & 66.95 & 1.4881 \\
        \modelname{}
        & $5\times10^{-7}$
        & $1\times10^{-3}$
        & $1\times10^{-3}$
        & 99.88
        & 98.51
        & 97.28
        & 100.00
        & 100.00
        & 85.53
        & 1.8787
        & 97.16
        & 97.16
        & 0.7766
        & 43.86
        & 0.9451 \\
        \midrule
        \multicolumn{16}{l}{\textbf{KL loss ablation}} \\

        KL
        & $1\times10^{-4}$ & -- & --
        & 34.78 & 0.00 & 0.00 & -- & --
        & -- & -- & 0.00 & 0.00
        & -- & -- & -- \\

        KL
        & $1\times10^{-5}$ & -- & --
        & $99.82$ & $96.35$ & $93.40$ & $100.00$ & $99.68$
        & $110.58$ & $2.4482$ & $93.20$ & $93.20$
        & $0.9301$ & $70.28$ & $1.5664$ \\

        KL
        & $1\times10^{-6}$ & -- & --
        & $99.68$ & $96.63$ & $95.00$ & $100.00$ & $99.65$
        & $107.43$ & $2.4040$ & $94.93$ & $94.93$
        & $0.9395$ & $64.60$ & $1.4515$ \\

        KL
        & $5\times10^{-7}$ & -- & --
        & 99.70 & 98.01 & 96.52 & 100.00 & 100.00
        & 90.66 & 2.0498 & 96.32 & 96.32
        & 0.7780 & 43.38 & 0.9852 \\

        KL
        & $1\times10^{-8}$ & -- & --
        & 99.70 & 97.22 & 95.53 & 100.00 & 100.00
        & 93.04 & 2.1498 & 95.53 & 95.53
        & 0.8564 & 47.00 & 1.0615 \\

        \midrule
        \multicolumn{16}{l}{\textbf{Perceptual loss ablation under $\lambda_{\mathrm{KL}}=5\times10^{-7}$}} \\

        KL + Perc.
        & $5\times10^{-7}$ & $5\times10^{-4}$ & --
        & 99.96 & 98.60 & 94.20 & 100.00 & 99.58
        & 79.29 & 2.4552 & 94.20 & 94.20
        & 0.8303 & 45.00 & 1.6775 \\

        KL + Perc.
        & $5\times10^{-7}$ & $7.5\times10^{-4}$ & --
        & 99.77 & 98.30 & 96.00 & 100.00 & 99.79
        & 80.18 & 1.8298 & 95.90 & 95.90
        & 0.8455 & 42.55 & 0.9856 \\

        KL + Perc.
        & $5\times10^{-7}$ & $1\times10^{-3}$ & --
        & 99.73 & 98.70 & 96.60 & 100.00 & 99.79
        & 82.63 & 1.8410 & 96.50 & 96.50
        & 0.8317 & 39.98 & 0.8765 \\

        KL + Perc.
        & $5\times10^{-7}$ & $5\times10^{-3}$ & --
        & 99.89 & 98.00 & 93.80 & 100.00 & 99.89
        & 87.93 & 1.9620 & 93.80 & 93.80
        & 0.9572 & 40.79 & 0.9178 \\

        KL + Perc.
        & $5\times10^{-7}$ & $1\times10^{-2}$ & --
        &  99.95 & 98.80 & 96.50 & 100.00 & 99.79
        & 91.17 & 2.0296 & 96.50 & 96.50
        & 0.8303 & 44.79 & 1.0008 \\

        \midrule
        \multicolumn{16}{l}{\textbf{REPA loss ablation under $\lambda_{\mathrm{KL}}=5\times10^{-7}$}} \\

        KL + REPA
        & $5\times10^{-7}$ & -- & $1\times10^{-6}$
        & 99.49 & 98.10 & 96.60 & 100.00 & 99.59
        & 102.86 & 2.2461 & 96.50 & 96.50
        & 0.8727 & 59.36 & 1.2727 \\

        KL + REPA
        & $5\times10^{-7}$ & -- & $1\times10^{-5}$
        & 99.39 & 97.70 & 96.00 & 100.00 & 99.48
        & 80.02 & 1.8149 & 96.00 & 96.00
        & 0.8181 & 37.61 & 0.8434 \\

        KL + REPA
        & $5\times10^{-7}$ & -- & $1\times10^{-4}$
        & 99.92 & 99.00 & 96.30 & 100.00 & 100.00
        & 90.84 & 2.0476 & 96.30 & 96.30
        & 0.9059 & 46.69 & 1.0501 \\

        KL + REPA
        & $5\times10^{-7}$ & -- & $1\times10^{-3}$
        & 99.74 & 98.00 & 96.40 & 100.00 & 99.69
        & 86.87 & 1.9782 & 96.30 & 96.30
        & 0.8091 & 45.83 & 1.0364 \\

        KL + REPA
        & $5\times10^{-7}$ & -- & $1\times10^{-2}$
        & 99.77 & 97.10 & 95.80 & 100.00 & 99.79
        & 86.29 & 1.9443 & 95.60 & 95.60
        & 0.8465 & 44.56 & 1.0059 \\

        \bottomrule
    \end{tabular}
    }
\end{sidewaystable}

\subsection{Transfer to an EDM Generator Backbone}
\label{app:edm-backbone}

We further evaluate \modelname{} with EDM~\citep{hoogeboom2022equivariant} as a second generator backbone in the representation-conditioned two-stage framework of GeoRCG~\citep{li2024geometric}. As shown in \Cref{tab:edm-backbone}, \modelname{} increases validity from $95.6\%$ to $97.5\%$ and molecule stability from $87.6\%$ to $94.2\%$, gains of $1.9$ and $6.6$ percentage points, respectively. Atom stability increases from $98.9\%$ to $99.3\%$, and uniqueness remains $100.0\%$.

\begin{table}[htbp]
    \centering
    \caption{Transfer of \modelname{} to an EDM generator backbone within the GeoRCG two-stage framework. All metrics are reported as percentages.}
    \label{tab:edm-backbone}
    \resizebox{0.94\linewidth}{!}{%
    \begin{tabular}{lcccc}
        \toprule
        Method & Validity $\uparrow$ & Uniqueness $\uparrow$ & Atom Stab. $\uparrow$ & Mol. Stab. $\uparrow$ \\
        \midrule
        GeoRCG-EDM baseline & 95.6 & 100.0 & 98.9 & 87.6 \\
        EDM + \modelname{} & \textbf{97.5} & 100.0 & \textbf{99.3} & \textbf{94.2} \\
        \bottomrule
    \end{tabular}%
    }
\end{table}

\subsection{Effect of Reducing Generation Steps}
\label{app:reduced_generation_steps}

A practical advantage of molecular generative models is their ability to generate high-quality samples with fewer sampling steps.
To evaluate whether \modelname{} improves sampling efficiency, we compare SemlaFlow and \modelname{} under different numbers of generation steps on the GEOM-DRUG dataset.
Both methods use the same generator architecture and sampling protocol, while \modelname{} replaces the original latent representation with the representation-head-refined latent.

Table~\ref{tab:vae_steps_compare} shows that from 20 sampling steps onward, \modelname{} improves all four metrics over SemlaFlow.
In particular, with only 20 sampling steps, \modelname{} achieves a validity of $0.967$, a connected-validity of $0.928$, and a molecule stability of $0.966$, already matching or exceeding the performance of SemlaFlow at 50 or 100 steps.
This suggests that the representation-head-refined latent provides a more effective conditioning signal, allowing the generator to reach high-quality molecular samples with fewer refinement steps.

The advantage is also reflected in energy-related quality.
At 10 steps, validity and molecule stability are on par with SemlaFlow, while energy per atom drops from $4.58$ to $3.93$. Across 10, 20, 50, and 100 sampling steps, \modelname{} obtains lower energy than SemlaFlow, indicating that the generated conformations are not only chemically valid but also more physically plausible.
For example, at 20 steps, \modelname{} reduces the energy from $2.30$ to $1.98$, and at 100 steps, from $2.06$ to $1.77$.

At 5 steps, neither method has converged. Overall, these results indicate that \modelname{} improves the sampling efficiency of SemlaFlow by enabling strong generation performance with substantially fewer generation steps.

\begin{table}[h]
    \centering
    \small
    \setlength{\tabcolsep}{5pt}
    \caption{Generation performance comparison between the baseline latent and the representation-head-refined latent under different sampling steps.}
    \label{tab:vae_steps_compare}
    \begin{tabular}{c|cc|cc|cc|cc}
    \toprule
    \multirow{2}{*}{Steps} & \multicolumn{2}{c|}{Validity} & \multicolumn{2}{c|}{Conn-Valid} & \multicolumn{2}{c|}{Mol-Stab} & \multicolumn{2}{c}{Energy/Atom} \\
     & \scriptsize SemlaFlow & \scriptsize \modelname{} & \scriptsize SemlaFlow & \scriptsize \modelname{} & \scriptsize SemlaFlow & \scriptsize \modelname{} & \scriptsize SemlaFlow & \scriptsize \modelname{} \\
    \midrule
    5   & 0.127 & \textbf{0.141} & \textbf{0.062} & 0.038 & \textbf{0.039} & 0.031 & 22.65 & \textbf{22.57} \\
    10  & \textbf{0.778} & 0.777 & \textbf{0.742} & 0.721 & 0.732 & \textbf{0.735} & 4.58 & \textbf{3.93} \\
    20  & 0.932 & \textbf{0.967} & 0.891 & \textbf{0.928} & 0.952 & \textbf{0.966} & 2.30 & \textbf{1.98} \\
    50  & 0.951 & \textbf{0.974} & 0.927 & \textbf{0.950} & 0.976 & \textbf{0.984} & 2.26 & \textbf{1.84} \\
    100 & 0.951 & \textbf{0.973} & 0.932 & \textbf{0.949} & 0.975 & \textbf{0.983} & 2.06 & \textbf{1.77} \\
    \bottomrule
    \end{tabular}
\end{table}

\subsection{Dataset Splits and Sampling Strategy}
\label{sec:dataset-splits}

For experiments on the \textbf{GEOM-DRUG} benchmark, we use both the full dataset and a representative subset. The full dataset, denoted as \emph{Fullset}, is split into 240,990 training molecules, 30,434 validation molecules, and 30,434 test molecules.

To enable efficient hyperparameter tuning and ablation studies under reduced computational cost, we additionally construct a representative subset, denoted as \emph{Subset}. The Subset contains 24,099 training molecules, corresponding to exactly 10\% of the full training split, and 6,087 validation molecules, corresponding to approximately 20\% of the full validation split. Importantly, the test set is kept identical to that of the Fullset, containing 30,434 molecules, for all subset experiments. This design ensures that evaluation metrics obtained from subset-trained models remain directly comparable to those of fullset-trained models and external baselines.

For \textbf{QM9} experiments, we adopt the commonly used split setting in prior 3D molecular generation work. The full QM9 dataset contains approximately 134k molecules. Following standard practice, we first randomly shuffle the dataset with a fixed random seed and assign 100,000 molecules to the training split. We then use 13,083 molecules as the test split and the remaining 17,748 molecules as the validation split. All QM9 experiments use the same split to ensure that different model variants are evaluated under identical data protocols.

\paragraph{Stratified Sampling by Atom Count.}
To ensure that the Subset preserves the structural complexity of the original chemical space, we adopt a stratified sampling strategy based on molecular size, defined by the exact number of atoms in each molecule. Preserving the atom-count distribution is important for training 3D molecular generative models, since both conformational diversity and generation difficulty are strongly affected by molecular size.

The subset construction is performed as follows. First, all molecules in each full split are grouped into distinct strata according to their integer atom counts. For each stratum, we compute its target number of selected samples according to its proportional frequency in the corresponding full split, and take the floor of this value as the initial allocation. Since this flooring operation may lead to a total count smaller than the desired subset size, we distribute the remaining sample slots using the largest-remainder method: strata with the largest fractional remainders receive one additional sample sequentially until the exact target subset size is reached. Within each stratum, molecular indices are randomly shuffled, and the allocated number of indices is selected. Finally, all selected indices from different strata are merged and globally shuffled to remove any ordering bias introduced by the stratification procedure.

\paragraph{Reproducibility.}
To make the subset splits exactly reproducible across different training runs and baseline evaluations, all random operations in intra-stratum selection and global shuffling are controlled by fixed random seeds. Specifically, we use seed 42 for constructing the training subset and seed 43 for constructing the validation subset.
\subsection{Computational Resources}
\label{app:computational_resources}

We report the computational resources used for training the molecular generator and the representation head in our main experiments.
All comparisons between the baseline generator and \modelname{} were conducted under the same training protocol and hyperparameter configuration, ensuring a fair assessment of the additional computational overhead introduced by our method.

On the GEOM-DRUG dataset, both the baseline and \modelname{} were trained for 300 epochs on Tesla A100 80GB GPUs with identical hyperparameters. \modelname{} adds about $1$GB of GPU memory and $30$ GPU hours, a $19\%$ increase in training cost (\Cref{tab:computational_resources}). The backbone architecture and the main generator training configuration remain unchanged.

On the QM9 dataset, both methods triggered early stopping within 8 GPU hours, and \modelname{} used less GPU memory than the baseline ($7.0$GB vs.\ $7.8$GB).

\begin{table}[h]
\centering
\caption{
Computational resources for training the molecular generator and representation head.
All GEOM-DRUG experiments were trained for 300 epochs on Tesla A100 80GB GPUs using identical hyperparameters for the baseline and \modelname{}.
On QM9, both methods triggered early stopping within 8 hours.
}
\label{tab:computational_resources}
\begin{tabular}{lcccc}
\toprule
Dataset & Method & GPU & GPU Memory & Training Cost \\
\midrule
GEOM-DRUG & Baseline & Tesla A100 80GB & 18.7GB & 155 GPU hours \\
GEOM-DRUG & \modelname{} & Tesla A100 80GB & 19.7GB & 185 GPU hours \\
\midrule
QM9 & Baseline & Tesla A100 80GB & 7.8GB & $<8$ GPU hours \\
QM9 & \modelname{} & Tesla A100 80GB & 7.0GB & $<8$ GPU hours \\
\bottomrule
\end{tabular}
\end{table}

\subsection{Baselines}
\label{appendix:baselines}

We compare against the following generators.

\textbf{MiDi}~\citep{midi} jointly diffuses the molecular graph and 3D coordinates: coordinates receive Gaussian noise in the zero center-of-mass subspace, while atom types, charges, and bond types follow categorical diffusion. The denoiser is a graph Transformer with relaxed EGNN layers, and an adaptive schedule noises atom types and charges faster than coordinates and bonds.

\textbf{JODO}~\citep{jodo} also models 2D topology and 3D geometry jointly. Its Diffusion Graph Transformer uses relational attention to update node and edge features together, and it supports conditioning on quantum properties.

\textbf{EQGAT-diff}~\citep{eqgatdiff} is an $E(3)$-equivariant attention-based diffusion model with $x_0$-parameterization and truncated-SNR time weighting. It adds hybridization, aromaticity, and ring features as inputs.

\textbf{SemlaFlow}~\citep{irwin2024semlaflow} replaces diffusion with equivariant flow matching and optimal-transport alignment, using latent graph attention (Semla) for message passing. It samples in as few as 20 steps. It is the backbone of GeoRCG and of \modelname{}.

\textbf{FlowMol}~\citep{flowmol} applies continuous flow matching to coordinates and continuous relaxations to categorical variables; its final version uses a Gaussian prior for categorical features rather than simplex-constrained flows.

\textbf{CanonFlow}~\citep{zhou2026rethinking} canonicalizes atom order and pose with spectral features so that a non-equivariant backbone can be trained on a single canonical slice; invariance is restored at sampling time by random symmetry transformations.

\textbf{GeoRCG}~\citep{li2024geometric} is the closest baseline to our setting. It first samples a molecule-level representation from a pretrained encoder's distribution with a representation diffusion model, then conditions a base generator (EDM or SemlaFlow) on that representation. \modelname{} builds on this framework (\Cref{sec:exp_setup}).

\section{Broader Impacts}
\label{sec:broader-impacts}

This work studies representation-conditioned 3D molecular generation, with the goal of improving the validity, stability, and physical plausibility of generated molecular structures. Its potential positive societal impacts include accelerating early-stage molecular design, reducing the cost of computational screening, and providing more reliable generative tools for scientific discovery. By improving the connection between learned molecular representations and 3D structure generation, the proposed method may contribute to more efficient exploration of chemical space and support applications in drug discovery, materials design, and molecular property optimization.

At the same time, molecular generative models may carry potential negative societal impacts. In particular, methods that improve the quality or efficiency of molecule generation could, in principle, be misused to propose harmful, toxic, or otherwise unsafe compounds. Although our work is foundational and does not directly optimize for toxicity, bioactivity, synthesizability, or deployment in real-world discovery pipelines, we acknowledge that improved molecular generation techniques may become part of broader systems with dual-use risks.

Several mitigation strategies are therefore important. Generated molecules should not be treated as actionable candidates without expert review, chemical validity checks, toxicity and safety screening, synthesizability assessment, and domain-specific experimental validation. In addition, practical deployment of such systems should consider access control, monitoring, and integration with filters for known toxicophores or restricted chemical classes. We release the method as a research contribution intended to improve scientific understanding of representation-conditioned molecular generation, rather than as an autonomous molecular design system.
\clearpage

\newtheorem{lenseslemma}{Lemma}[section]
\newtheorem{lensesproposition}[lenseslemma]{Proposition}
\newtheorem{lensescorollary}[lenseslemma]{Corollary}
\crefname{lenseslemma}{Lemma}{Lemmas}
\Crefname{lenseslemma}{Lemma}{Lemmas}
\crefname{lensesproposition}{Proposition}{Propositions}
\Crefname{lensesproposition}{Proposition}{Propositions}
\crefname{lensescorollary}{Corollary}{Corollaries}
\Crefname{lensescorollary}{Corollary}{Corollaries}
\newenvironment{lensesproof}{\par\noindent\textit{Proof.}\ }{\hfill$\square$\par\medskip}

\section{Theoretical Motivation for Representation Refinement}
\label{app:lenses-theory}

This section gives statements and proofs for the theoretical motivation behind the three representation-refinement objectives. We treat four aspects separately---latent regularity, conditional feature readout, feature-distribution discrepancy, and a local optimization surrogate---since each rests on different assumptions.

\subsection{Definitions and Scope}
\label{app:theory-scope}

Let $(\mathcal X,d_{\mathrm{mol}})$ be a molecular data space, and let $\boldsymbol\mu_\psi:\mathcal X\to\mathbb R^{d_r}$ be the mean produced by the learned representation head on top of the frozen encoder. In a fixed-composition local region, $d_{\mathrm{mol}}$ can be defined on molecular configurations modulo the relevant rigid-motion and atom-permutation symmetries.

The Phase-I stochastic condition is $\bm z=\boldsymbol\mu_\psi(\mathcal M)+\boldsymbol\sigma_\psi(\mathcal M)\odot\bm\epsilon$, whereas the Phase-II RDM targets the distribution of the deterministic mean $R_0=\boldsymbol\mu_\psi(\mathcal M)$. We use Euclidean norms in representation space, operator norms for matrices, and expectations over the explicitly stated data, noise, or prediction--target coupling. The encoder and representation-head parameters are fixed when analyzing a learned feature map.

\paragraph{Scope.} The results below are heuristic motivations derived under idealized assumptions (Lipschitz or bi-Lipschitz feature maps, bounded support, fixed learned maps, and a local quadratic model), not guarantees for the trained model. They explain why each objective is expected to help; the evidence that it does help in practice is the ablation study in \Cref{ablation} and \Cref{app:fullset_ablations}.

\subsection{Noisy Latent-Score Regularity}
\label{app:score-regularity}

\begin{lenseslemma}[Gaussian smoothing of a bounded mean-space representation]
\label{lem:lenses-score}
Suppose the support of the molecular data distribution has diameter at most $D_X$ in $d_{\mathrm{mol}}$, and $\boldsymbol\mu_\psi$ is $L_\mu$-Lipschitz on that support. Let
\begin{equation}
R_t=\alpha_tR_0+\sigma_t\xi,\qquad
R_0=\boldsymbol\mu_\psi(\mathcal M),\qquad
\xi\sim\mathcal N(0,I),\quad\sigma_t>0.
\end{equation}
If $q_t$ is the density of $R_t$, then
\begin{equation}
\operatorname{Lip}(\nabla\log q_t)
\le\sigma_t^{-2}
+\frac{\alpha_t^2L_\mu^2D_X^2}{4\sigma_t^4}.
\label{eq:lenses-score-bound}
\end{equation}
\end{lenseslemma}

\begin{lensesproof}
Write $Y=\alpha_tR_0$ and $\sigma=\sigma_t$. The density is the Gaussian convolution
\begin{equation}
q_t(x)=\int\phi_\sigma(x-y)\,dP_Y(y).
\end{equation}
Bounded support permits differentiation under the integral. With $m(x)=\mathbb E[Y\mid R_t=x]$, differentiating the Gaussian kernel gives
\begin{equation}
\nabla\log q_t(x)=\sigma^{-2}(m(x)-x).
\end{equation}
The derivative of the log posterior density with respect to $x_j$ is $(y_j-m_j(x))/\sigma^2$. Hence
\begin{equation}
\frac{\partial m_i(x)}{\partial x_j}
=\sigma^{-2}\operatorname{Cov}(Y_i,Y_j\mid R_t=x),
\end{equation}
and therefore
\begin{equation}
\nabla^2\log q_t(x)
=-\sigma^{-2}I+\sigma^{-4}\operatorname{Cov}(Y\mid R_t=x).
\label{eq:lenses-score-hessian}
\end{equation}
The support of $Y$ has diameter at most $D_Y=|\alpha_t|L_\mu D_X$. For any unit vector $v$, the scalar $v^TY$ lies in an interval of length at most $D_Y$. If $c_v$ is the midpoint of this interval, then $\operatorname{Var}(v^TY\mid R_t=x)\le\mathbb E[(v^TY-c_v)^2\mid R_t=x]\le D_Y^2/4$. Thus
\begin{equation}
\|\operatorname{Cov}(Y\mid R_t=x)\|_{\mathrm{op}}\le D_Y^2/4.
\end{equation}
Taking operator norms in \Cref{eq:lenses-score-hessian} yields the claimed bound uniformly in $x$. Integrating the bounded Hessian along a line segment proves the global Lipschitz bound for the score.
\end{lensesproof}

For a fixed noise schedule and molecular support, a smaller $L_\mu$ tightens the second term of \Cref{eq:lenses-score-bound}, so a smoother representation map yields a better-conditioned latent score at every positive noise level.

\paragraph{Relation to Representation Sampling Bounds.}
GeoRCG Proposition D.4~\citep{li2024geometric} depends on score regularity, a representation second-moment bound, score-estimation error, and a compatible discretization step size. Lemma~\ref{lem:lenses-score} provides a regularity estimate at positive noise levels. On an interval $[\tau,T]$ with $\tau>0$, a schedule with $\inf_{t\in[\tau,T]}\sigma_t>0$ and bounded $\alpha_t$ gives a finite uniform score bound, which can be combined with an early-stopping argument controlling the bias between $q_\tau$ and $q_0$.

\subsection{KL Regularization and Latent Second Moments}
\label{app:kl-moment}

\begin{lensesproposition}[Posterior KL controls the mean-space second moment]
\label[lensesproposition]{prop:lenses-kl}
Let $q_\psi(z\mid\mathcal M)=\mathcal N(\boldsymbol\mu_\psi(\mathcal M),\operatorname{diag}(s_1^2,\ldots,s_{d_r}^2))$, with every $s_j^2>0$, and let $\gamma=\mathcal N(0,I)$. For the summed Gaussian posterior KL used in \Cref{eq:kl-detail},
\begin{equation}
\mathbb E\|R_0-\mathbb ER_0\|^2
\le\mathbb E\|R_0\|^2
\le2\mathbb E_{\mathcal M}\operatorname{KL}(q_\psi(\cdot\mid\mathcal M)\|\gamma).
\label{eq:lenses-kl-moment}
\end{equation}
\end{lensesproposition}

\begin{lensesproof}
The Gaussian KL formula gives
\begin{equation}
\operatorname{KL}(q_\psi(\cdot\mid\mathcal M)\|\gamma)
=\frac12\sum_{j=1}^{d_r}\left(\mu_j^2+s_j^2-1-\log s_j^2\right).
\end{equation}
Since $a-1-\log a\ge0$ for $a>0$, the expression is at least $\|\boldsymbol\mu_\psi(\mathcal M)\|^2/2$. Taking expectations proves the second inequality. The first follows from $\mathbb E\|R_0-\mathbb ER_0\|^2=\mathbb E\|R_0\|^2-\|\mathbb ER_0\|^2$.
\end{lensesproof}

The posterior KL, the aggregated stochastic posterior $q_z=\mathbb E_{\mathcal M}q_\psi(z\mid\mathcal M)$, and the mean-space distribution $q_\mu=(\boldsymbol\mu_\psi)_\#P_{\mathcal M}$ must be distinguished. Convexity gives
\begin{equation}
\operatorname{KL}(q_z\|\gamma)
\le\mathbb E_{\mathcal M}\operatorname{KL}(q_\psi(\cdot\mid\mathcal M)\|\gamma),
\end{equation}
whereas for the deterministic-mean distribution $q_\mu$ the relevant controlled quantity is the second moment in \Cref{eq:lenses-kl-moment}, which is the term that enters representation-sampling bounds. When the KL is averaged rather than summed over latent coordinates, the right-hand side of \Cref{eq:lenses-kl-moment} is multiplied by $d_r$.

\subsection{Representation Alignment and Conditional Prediction}
\label{app:repa-readout}

\begin{lensesproposition}[An oracle-readout bound for normalized alignment]
\label[lensesproposition]{prop:lenses-repa}
Let $u$ and $e$ be unit-norm projected generator and encoder features, and define $\ell_{\mathrm{align}}=1-\langle u,e\rangle$. Suppose an oracle target $s^*$ admits an $L_A$-Lipschitz feature readout $A$ with $\mathbb E\|A(e)-s^*\|^2\le\rho^2$. Then
\begin{equation}
\mathbb E\|A(u)-s^*\|^2
\le4L_A^2\mathbb E\ell_{\mathrm{align}}+2\rho^2.
\label{eq:lenses-repa-oracle}
\end{equation}
If the actual predictor $s_\theta$ has realization residual $\delta^2=\mathbb E\|s_\theta-A(u)\|^2$, then
\begin{equation}
\mathbb E\|s_\theta-s^*\|^2
\le3\delta^2+6L_A^2\mathbb E\ell_{\mathrm{align}}+3\rho^2.
\label{eq:lenses-repa-realization}
\end{equation}
\end{lensesproposition}

\begin{lensesproof}
Unit normalization gives $\|u-e\|^2=2(1-\langle u,e\rangle)$. Split $A(u)-s^*=[A(u)-A(e)]+[A(e)-s^*]$, apply $\|a+b\|^2\le2\|a\|^2+2\|b\|^2$, and use the Lipschitz condition to obtain \Cref{eq:lenses-repa-oracle}. For the actual predictor, use the three-term decomposition
\begin{equation}
s_\theta-s^*=[s_\theta-A(u)]+[A(u)-A(e)]+[A(e)-s^*].
\end{equation}
The inequality $\|a+b+c\|^2\le3(\|a\|^2+\|b\|^2+\|c\|^2)$ gives \Cref{eq:lenses-repa-realization}.
\end{lensesproof}

The readout can additionally depend on the noisy state, time, and conditioning representation, provided its Lipschitz constant in the aligned feature is uniform over the considered setting. The bound shows that, for a fixed readout, the alignment loss directly controls the only term in the conditional prediction error that depends on the generator's intermediate features.

The implemented REPA objective is negative cosine similarity, whereas the discrepancy in the proposition is nonnegative $1-\cos$. Under the same averaging convention, $\ell_{\mathrm{align}}=1+\mathcal L_{\mathrm{REPA}}$. For node-wise losses, the proposition applies directly to a node-level target; a molecular-level readout picks up the corresponding atom-count or averaging factor.

\subsection{Perceptual Loss and Feature-Distribution Discrepancy}
\label{app:perceptual-wasserstein}

\begin{lenseslemma}[Paired feature error bounds Wasserstein discrepancy]
\label{lem:lenses-perceptual}
Let $(\widehat{\mathcal M},\mathcal M)$ have the joint distribution of a denoising prediction and its clean target. Fix the pooled feature map $g$, and assume its two induced distributions have finite second moments. Then
\begin{equation}
W_2^2\left(g_\#P_{\widehat{\mathcal M}},g_\#P_{\mathcal M}\right)
\le\mathbb E\|g(\widehat{\mathcal M})-g(\mathcal M)\|^2.
\label{eq:lenses-feature-wasserstein}
\end{equation}
If the time weighting satisfies $w(t)\ge w_{\min}>0$ over the considered time distribution, the right-hand side is at most $\mathbb E\mathcal L_{\mathrm{PERC}}/w_{\min}$.
\end{lenseslemma}

\begin{lensesproof}
Push the prediction--target coupling forward through $(g,g)$. The result is a valid coupling of the two feature distributions. Since $W_2^2$ is the infimum of the expected squared distance over all such couplings, it is no greater than the cost of this particular pairing. The weighted statement follows pointwise from $w_{\min}\|g(\widehat{\mathcal M})-g(\mathcal M)\|^2\le\mathcal L_{\mathrm{PERC}}$ and then taking expectations.
\end{lensesproof}

\begin{lensescorollary}[Geometric error under local non-collapse]
\label[lensescorollary]{cor:lenses-geometry}
If, for the paired molecules in a specified local region, $g$ additionally satisfies
\begin{equation}
\|g(\mathcal M)-g(\mathcal M')\|\ge c\,d_{\mathrm{mol}}(\mathcal M,\mathcal M'),\qquad c>0,
\end{equation}
then
\begin{equation}
\mathbb E d_{\mathrm{mol}}(\widehat{\mathcal M},\mathcal M)^2
\le c^{-2}\mathbb E\|g(\widehat{\mathcal M})-g(\mathcal M)\|^2.
\end{equation}
\end{lensescorollary}

\begin{lensesproof}
Square the assumed lower bound for each pair and take expectations.
\end{lensesproof}

The corollary relies on a lower (non-collapse) bound on $g$ rather than an upper Lipschitz bound, which is why the effective-rank and pairwise-distance diagnostics in \Cref{sec:appendix-pairwise,sec:appendix-rep-quality} are relevant: they measure whether the refined feature map avoids collapse.

\subsection{A Local Quadratic Optimization Surrogate}
\label{app:quadratic-surrogate}

\begin{lensesproposition}[Isotropic regularization of a quadratic surrogate]
\label[lensesproposition]{prop:lenses-conditioning}
Consider direct optimization of a feature vector $u$ with local quadratic objective $F(u)=\tfrac12(u-u_*)^TH(u-u_*)$, where $H$ is positive definite. Let $m=\lambda_{\min}(H)>0$ and $L=\lambda_{\max}(H)$. Adding $\tfrac\lambda2\|u-e\|^2$, with a fixed target $e$ and $\lambda\ge0$, changes the Hessian to $H+\lambda I$ and gives
\begin{equation}
\kappa(H+\lambda I)=\frac{L+\lambda}{m+\lambda}\le\frac Lm=\kappa(H).
\label{eq:lenses-conditioning}
\end{equation}
\end{lensesproposition}

\begin{lensesproof}
Each eigenvalue increases by $\lambda$. The inequality follows from $m(L+\lambda)\le L(m+\lambda)$. It is strict when $L>m$ and $\lambda>0$.
\end{lensesproof}

The proposition is stated in feature space. In parameter space, with $J=\nabla_\theta u_\theta$, the alignment term contributes $\lambda J^T J$ to the Gauss--Newton Hessian, so the same isotropic effect holds to the extent that $J^T J$ is well conditioned on the relevant subspace.

\end{document}